\documentclass[journal]{IEEEtran}
\usepackage{graphicx}
\usepackage{ifpdf}
\usepackage{cite}
\usepackage{amsmath}
\usepackage{amssymb}
\usepackage{bbm}
\usepackage{graphicx}
\usepackage{subcaption}
\usepackage{siunitx}
\usepackage{algorithmic}
\usepackage{array}
\usepackage[dvipsnames]{xcolor}

\usepackage{stfloats}
\usepackage{url}
\usepackage[switch]{lineno}
\usepackage{color}
\usepackage[colorlinks=true,linkcolor=red, citecolor=green, urlcolor=blue]{hyperref}
\usepackage{rotating}
\usepackage{gensymb}
\usepackage{textcomp}
\usepackage{booktabs,array,dcolumn}
\usepackage{graphicx} 
\usepackage{multirow} 
\usepackage{array, multirow}
\usepackage{booktabs}
\usepackage{tabularx}
\usepackage{graphicx}
\usepackage{array,makecell}
\usepackage{multirow}

\begin{document}

\title{An Autoencoder Architecture for L-band Passive Microwave Retrieval of Landscape Freeze-Thaw Cycle}

\author{Divya Kumawat, Ardeshir Ebtehaj, Xiaolan Xu, Andreas Colliander, and Vipin Kumar 
\thanks{D. Kumawat, and A. Ebtehaj are with the Department of Civil Environmental and Geo- Engineering and the Saint Anthony Falls Laboratory, University of Minnesota, Minneapolis, MN 55455 USA. e-mail: (ebtehaj@umn.edu). X.Xu and A.Colliander are with the NASA Jet Propulsion Laboratory, California Institute of Technology, Pasadena, CA 91109 USA. V.Kumar is with the University of Minnesota, Minneapolis, MN 55455}}

\markboth{IEEE Transactions on Geoscience and Remote Sensing}
{Divya \MakeLowercase{\textit{et al.}}: Deep Learning of the Soil Freeze-Thaw State using SMAP Brightness Temperature}

\maketitle
\begin{abstract}

Estimating the landscape and soil freeze-thaw (FT) dynamics in the Northern Hemisphere is crucial for understanding permafrost response to global warming and changes in regional and global carbon budgets. A new framework is presented for surface FT-cycle retrievals using L-band microwave radiometry based on a deep convolutional autoencoder neural network. This framework defines the landscape FT-cycle retrieval as a time series anomaly detection problem considering the frozen states as normal and thawed states as anomalies. The autoencoder retrieves the FT-cycle probabilistically through supervised reconstruction of the brightness temperature (TB) time series using a contrastive loss function that minimizes (maximizes) the reconstruction error for the peak winter (summer). Using the data provided by the Soil Moisture Active Passive (SMAP) satellite, it is demonstrated that the framework learns to isolate the landscape FT states over different land surface types with varying complexities related to the radiometric characteristics of snow cover, lake-ice phenology, and vegetation canopy. The consistency of the retrievals is evaluated over Alaska, against in situ ground-based observations, showing reduced uncertainties compared to the traditional methods that use thresholding of the normalized polarization ratio.  

\end{abstract}

\begin{IEEEkeywords}
SMAP Satellite, deep learning, Convolutional autoencoders, Soil Freeze and Thaw, Snow, Snow wetness, L-band microwaves, Soil Remote Sensing
\end{IEEEkeywords}

\IEEEpeerreviewmaketitle

\section{Introduction}

\IEEEPARstart{F}{reez}-thaw (FT) seasonal cycles occur over more than 50\% of terrestrial landscapes in the Northern Hemisphere (NH) \cite{zhang2003distribution} and exhibit high spatial and temporal variability -- affecting the energy, water, and carbon balance and the downstream biogeochemical \cite{schaefer2011amount}, ecological \cite{black2000increased}, and hydrological processes \cite{gouttevin2012insulating}. The ground's FT cycle is closely related to the soil's physical properties, vegetation, and snow cover. The timing and duration of the soil FT cycle affect infiltration of rainfall and snowmelt \cite{koren1999scale} and thus control the dynamics of nutrient availability and plant growth. As the Arctic and its boreal forests continue to warm, evidence suggests that the frequency of the FT cycle will change, making it a critical variable for predicting the impacts to the soil and vegetation carbon exchange with the atmosphere in a warming climate \cite{osterkamp2005recent, romanovsky2010permafrost}.

The capacity to perform direct sampling and on-site ground-based observations of the FT cycle is presently restricted and logistically constrained globally, especially in the high latitudes. Satellite microwave radiometry offers a cost-effective alternative \cite{wegmuller1990effect} in all weather conditions with no limitations concerning seasonal high-latitude darkness. This is attributed to the pronounced sensitivity of the landscape emissivity to the presence or absence of liquid water. The reason is that the relative permittivity of ice (i.e., $\epsilon' \approx 3.15$) and freshwater (i.e., $\epsilon' \approx 100-10$) are markedly different in microwave frequencies 1--50 GHz. When the landscape freezes up, the surface emissivity and the observed microwave brightness temperatures (TB) increase significantly during the winter. However, the emissivity drops rapidly during the thaw period, giving rise to colder radiometric temperatures that can vary appreciably over summer due to changes in soil moisture and vegetation water content. Currently, key satellites that can provide all-weather observations of surface emission under moderate vegetation water content include the L-band (1.4 GHz) NASA Soil Moisture Active Passive Mission (SMAP) \cite{entekhabi2010soil} and the European Space Agency's (ESA) Soil Moisture and Ocean Salinity (SMOS) \cite{kerr2010smos} mission. 

The current SMAP FT product uses the seasonal thresholding of the departure of the normalized polarization ratio (NPR) \cite{dunbar2016smap, xu2016landscape} from its reference FT values. The NPR is the ratio of the difference between TBs at horizontal and vertical polarization to their summation \cite{xu19681jet,xu2018global}. After thresholding the NPR, two additional steps are taken to reduce false characterization of the FT cycle. Firstly, if the TBs at both polarization values exceed 273~\si{K}, the pixel is marked as thawed regardless of the retrieval outcome. Second, ``never frozen'' and ``never thawed'' masks were used based on a long-term (2002-2019) daily FT cycle data set \cite{kim2012satellite}, produced from observations by the Advanced Microwave Scanning Radiometer (AMSR) and the simulated surface temperatures by the NASA’s Goddard Modeling and Assimilation Office \cite{derksen2017retrieving}.

The NPR algorithm is sensitive to the inherent variability of TB time series not only in response to changes in soil permittivity but also to the presence of water in snow, vegetation, and sub-grid lakes. Consequently, the optimal selection of reference NPR values and the thresholds remain a major source of uncertainty in the produced FT cycle data \cite{kim2019global}. For instance, it is found that establishing the reference values requires relatively stable frozen or thawed conditions with a minimum duration of 20 days \cite{dunbar2016smap}. This requirement is often a limiting factor for capturing high-frequency FT cycles. It is also shown that adopting a fixed global threshold seems insufficient to accommodate the diverse signals from structurally and radiometrically complex land surfaces -- characterized by varying soil textures, snow dynamics, and vegetation phenology \cite{lv2022novel}.

As is well understood, satellite observations of the upwelling brightness intensity respond to the presence or absence of liquid water in the field of view (FOV). When snow falls on the ground, the soil remains wet for an extended time depending on the air temperature, depth of the snowpack, and soil mineralogy \cite{watanabe2002amount, watanabe2009measurement, mavrovic2021soil, gao2022variability, kumawat2022passive, kumawat2024global, kumawat2023passive}. On the other hand, when the snowpack melts, the TBs respond to snow wetness and not necessarily the thawing soil \cite{naderpour2017davos, derksen2017retrieving}. Moreover, there is often a time lag between the freeze-up and break-up times of the lake ice and the FT cycle of the surrounding landscape \cite{bergstedt2020influence}, which can affect the NPR and the optimal thresholding approach in unknown ways. The impacts of the sub-grid lake water fraction on the TBs and NPR are not yet well understood and can be an important factor over the Arctic, where a large fraction of the FOV is populated with lakes of different sizes. Therefore, with these sources of uncertainties, it seems more realistic to characterize the FT cycle of the landscape probabilistically through a generalized machine-learning mechanism rather than empirical thresholding.

In recent years, modern machine learning frameworks have emerged to learn the landscape FT dynamics probabilistically using multi-frequency observations from L- to Ka-band. Both supervised and unsupervised methodologies have been adopted. The supervised approaches considered classic decision trees \cite{li2022machine,zhong2022freeze} and deep U-Net \cite{donahue2023deep} architectures. The characterization of the FT cycle was defined as a classification or an image segmentation problem without explicitly accounting for its widely used temporal microwave signatures \cite{roy2015evaluation}. The supervised approaches naturally require many labeled data, which can be a limitation over northern latitudes. Previous studies derived the labels from a combination of sparse ground-based observations and reanalysis datasets \cite{donahue2023deep}, which can make the results prone to model errors. 

Unsupervised methods can bypass the bottleneck of ground-based data for labeling. A Hidden Markov Model (HMM) was developed \cite{walker2022satellite} to estimate FT state probability using multi-frequency data from SMAP and AMSR radiometers. The model was trained based on the data points during the shoulder seasons and demonstrated marginal performance gain, compared to the SMAP operational product that relies on NPR thresholding. A source of uncertainty might arise as the training time series shall precisely represent the duration of the shoulder seasons, which is not precisely known a priori. Moreover, unsupervised methods can only learn the structure of the training data (e.g., shoulder seasons) and cannot distinguish them from those unseen data points(i.e., summer or winter seasons).

This study offers an alternative approach by recasting the landscape FT retrieval as a time series anomaly detection problem. In time series analysis, an anomaly represents a temporal segment that statistically deviates from the expected or normal behavior of the time series \cite{choi2021deep, chandola2009anomaly}. Anomaly detection has been an important task with a wide range of applications for example in the detection of heart attacks \cite{ansari2017review}, structural defects \cite{woike2014structural}, or ecosystem disturbances in earth sciences \cite{cheng2009detection}. Various methods leveraging supervised \cite{shaukat2021review} and unsupervised \cite{kiran2018overview} deep learning techniques were proposed.

Conceptually, the proposed FT-Cycle autoencoder (FTC-Encoder) framework is consistent with the classic formulation of the FT cycle: An anomalous departure of the observed TBs from their reference frozen or thawed climatological values. However, unlike previous work, from a practical standpoint, the approach relies on (i) a supervised convolutional autoencoder architecture that incorporates the temporal information of the TB time series; (ii) produces a probability measure representing the state of the surface FT cycle, and (iii) only needs labels for frozen (peak winter) and thawed (peak summer) segments of time series, which can be provided with minimum uncertainty and almost no reliance on ground-based observations. 

The FTC-Encoder consists of a series of down-sampling convolution operations to produce a low-dimensional latent space representation of the SMAP TB time series followed by blocks of upsampling operations via transposed convolution operations. It learns the FT states from variable-length time series of microwave TBs, collected over different land-cover types with variable sub-grid water fractions. The network is trained based on a contrastive loss function \cite{yamanaka2019autoencoding} that minimizes a reconstruction mean-squared error for normal samples (i.e., frozen segments) while maximizing it for the known anomalous samples (i.e., thawed segments). This article is organized as follows: Section~\ref{sec:II} details the used datasets and Section~\ref{sec:III} elaborates on the algorithm and its implementation. Section~\ref{sec:IV} presents the results, while Section~\ref{sec:V} concludes our study, addressing current limitations and suggesting potential research extensions. 

\section{Data and Study Area}
\label{sec:II}

\subsection{SMAP data}
\label{sec:II.1}

The SMAP level-3 enhanced TBs \cite{Chan2018, oneill2021_l3smpe_product} at both polarization channels, gridded on the EASE (Equal-Area Scalable Earth) version 2  9~\si{km} grid \cite{brodzik2012ease2}, from the morning overpasses in calendar years 2015--2020 were used. The temporal resolution of the SMAP radiometer morning overpasses ranges from 1 to 2 days in the high-latitude study area; the revisit time is 2--3 days on the equator. The daily time series of TB values are obtained through linear interpolation on a per-grid cell basis. Ancillary information is also utilized including the International Geosphere-Biosphere Programme (IGBP) \cite{loveland2000development} land-cover type based on MODIS-MCD12QI product \cite{friedl2002global} and a water mask based on MODIS-MOD44W-v6 product \cite{carroll5others,carroll2017development}.

\begin{figure}
    \centering
    \includegraphics[width=0.35\textwidth]{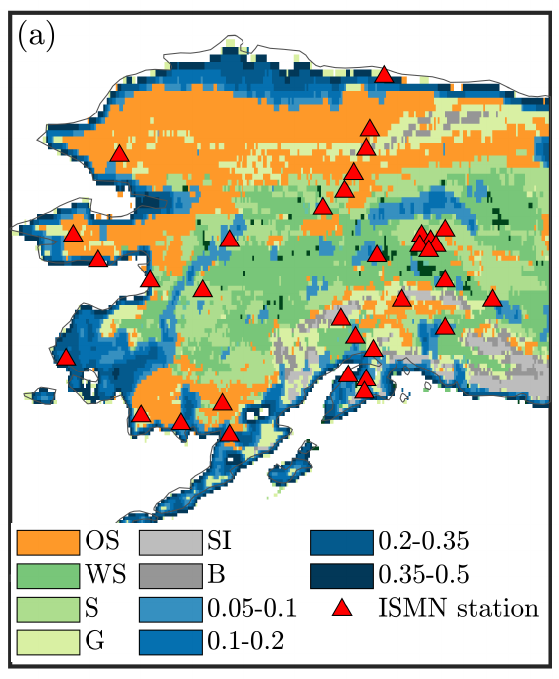}\\
    \includegraphics[width=0.47\textwidth]{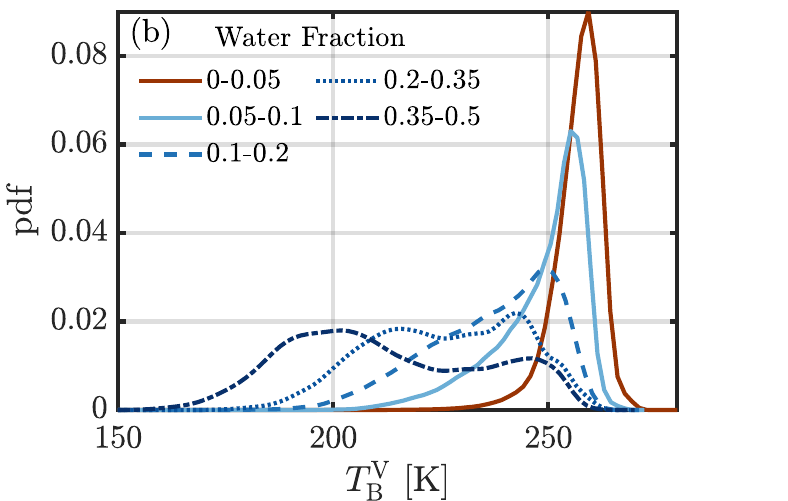}\\
    \caption{(a) The study area overlaid with IGBP land-cover types, water-fraction classes, the spatial distribution of 34 available ISMN stations; and (b) the probability density function (pdf) of SMAP TBs in 2018 at vertical polarization for all pixels with varying water fractions. The land-cover map represents open shrublands (OS), woody savannas (WS), savannas (S), grasslands (G), Permanent Ice/snow (SI), and Barren (B) for areas with water fraction less than 0.05.}
    \label{fig:01}
\end{figure}

The SMAP level-3 passive FT state data (SPL3FTP\_E) \cite{dunbar2016algorithm} is compared with the FTC-Encoder results. This product is a daily estimate of the FT state on the EASE version 2 9~\si{km} grid, over landscape above $45^\circ \rm {N}$ based on a seasonal threshold algorithm \cite{dunbar2016smap}. Here, the study area is confined to the state of Alaska (i.e., 60--72$^\circ$N, 140-170$^\circ$W) in the United States (Fig.~\ref{fig:01}a). The predominant land-cover types include open shrublands (OS), woody savannas (WS), savannas (S), and grasslands (G). The Alaskan landscape is populated with numerous inland water bodies, represented through surface classes with different sub-grid scale water fractions within the SMAP grids.

Figure~\ref{fig:01}b presents the probability density function (pdf) of annual TB values at vertical polarization for pixels with varying water fractions, in which the horizontal polarization is omitted for brevity. As water fractions increase, TB values decrease, and the shape of the distribution transitions from unimodal to bimodal. This shift can be attributed to a significant increase in TB values from summer to winter, and the effects of ice and snow-cover dynamics on the bulk surface emissivity dynamics.

\subsection{ERA5 land data sets}
\label{sec:II.2}

The gridded hourly and daily surface soil and air temperatures from the hourly ERA5 land reanalysis data \cite{hersbach2018copernicus,munoz2021era5} are used at a nominal resolution 9~\si{km}. The near-surface (i.e., 0--7 cm depth) soil temperatures at 6 a.m. and 2-meter air temperatures were used to label the SMAP TB time series for training purposes, as explained in Section~\ref{sec:III.2}. 

\subsection{In situ data} 
\label{sec:II.3}

To examine the quality of FT-cycle retrievals, we resort to the collection of {\it in situ} ground-based soil moisture networks by the International Soil Moisture Network (ISMN) \cite{dorigo2011international}. This collection contains measurements of more than 1,500 gauge stations from various international networks. We use the nearest 2-meter air and soil temperatures (top 5 cm) at the time of SMAP overpasses from stations available in three networks: Soil Climate Analysis Network (SCAN, 14 stations, \cite{schaefer2007usda}), snow telemetry (SNOTEL, 19 stations, \cite{leavesley2008modeling}), and the U.S. Climate Reference Network (USCRN, 1 station, \cite{bell2013us}), as shown in Fig.~\ref{fig:01}. Only the data with good-quality flags are used. The location of each station measurement was assigned to the nearest SMAP grid and averaged if multiple stations were positioned in one grid. The {\it in situ} air and ground temperature data are labeled frozen for values less than $0$~\si{\degreeCelsius}.

\section{Methodology}
\label{sec:III}

\subsection{Autoencoder Models and Loss Function}
\label{sec:III.1}

Autoencoders \cite{rumelhart1986parallel}, commonly adopted for unsupervised tasks, represent a specialized class of deep neural networks capable of encoding (downsampling) input data into a compact latent-space representation and then decoding (upsampling) it back to the original input space through minimization of a reconstruction loss function. For unsupervised learning, the loss function is typically organized as a mean-squared reconstruction error:
\begin{equation}
    \mathcal{L}_\theta(\mathbf{x}) = \Vert \mathbf{x}-D\left(E(\mathbf{x,\theta})\right)\Vert^2_2,
    \label{eq_01}
\end{equation}
where $\mathbf{x} = (x_1,x_2,\ldots,x_n)$ is the input time series, $E(\mathbf{\cdot})$ and $D(\cdot)$ are the encoder and decoder functions, $\Vert\cdot\Vert_2^2$ denotes $\ell_2$-norm, and $\theta$ represents a set of the trainable parameters. 

As previously noted, unsupervised autoencoders were used for anomaly detection tasks \cite{kiran2018overview,shaukat2021review}. Commonly, they are trained for normal data points and produce relatively large reconstruction errors for abnormal data points -- allowing anomaly detection \cite{chen2018autoencoder} through a statistical thresholding mechanism \cite{givnan2022anomaly,ndubuaku2019unsupervised}. For the FT-cycle problem, we can define the TB time series over the frozen winter as normal points expecting an autoencoder to produce large errors when the landscape is thawed (known anomalies), enabling FT-cycle retrievals. 

However, we found that such a problem formulation can be ineffective since the autoencoder exhibits a low discrimination power due to similarities between the latent space representation of frozen and thawed time series. As a result, an unsupervised autoencoder often produces low reconstruction errors even for time series over the peak summer. Furthermore, this paradigm still relies on a thresholding scheme that can be subjective and sub-optimal due to space-time variability of land surface structural and radiometric properties. Moreover, the time window when the landscape transitions between the fully frozen and thawed states (mixed anomalies) is unclear as a priori, exacerbating the error of FT-cycle retrievals. 

To address these limitations, our study proposes a new supervised convolutional autoencoder FT-Cycle retrieval framework, called FTC-Encoder, leading to a probabilistic representation of the FT cycle without any subjective thresholding. The main practical advantage is that this framework can be trained for labeled time series with variable lengths enabling supervised training over the peak summer and winter when we are certain about the landscape FT status.  

\begin{figure*}
    \centering
    \includegraphics[width=1\textwidth]{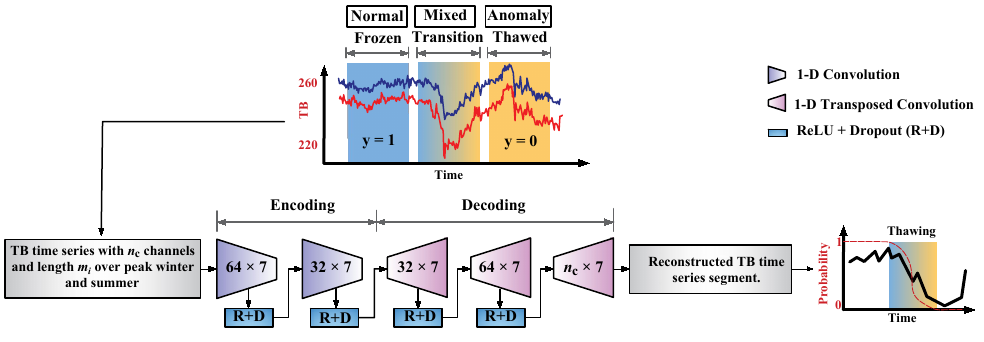}
    \caption{A schematic of the FTC-Encoder architecture: A supervised convolutional autoencoder framework that can learn from the peak winter and summer TB time series with variable lengths to probabilistically determine the FT state.}
    \label{fig:02}
\end{figure*}

Suppose that we have a training dataset $\{(\mathbf{x}_1,y_1), (\mathbf{x}_2,y_2),$ $\ldots, (\mathbf{x}_n,y_n)\}$, where $\mathbf{x}_i\in\mathbb{R}^{n_c\times m_i}$ represents the $i$-th time series at $n_c$ channels with variable length $m_i$, $y_i \in \{0,1\}$ denotes its labels that is $y_i = 1$ for a fully frozen normal data point and $y_i = 0$ for fully thawed anomalous data points. Here we assume that the conditional probability of FT status $y_i$ given a TB time series $\mathbf{x}_i$ follows the Bernoulli distribution: 

\begin{equation}
p(y_i|\mathbf{x}_i) = [\eta_\theta(\mathbf{x}_i)]^{y_i} [1-\eta_\theta(\mathbf{x}_i)]^{1-y_i},
\label{eq_02}
\end{equation}
where $\eta_\theta(\mathbf{x}_i) = e^{-\mathcal{L}_\theta(\mathbf{x}_i)}$. This assumption leads to the following negative log-likelihood function \cite{yamanaka2019autoencoding}:
\begin{align}
-&\log p_\theta(y_i|\mathbf{x}_i)\notag \\
&= -y_i \log \eta_\theta(\mathbf{x}_i) - (1-y_i)\log\left[1-\eta_\theta(\mathbf{x}_i)\right]\notag \\
     &= y_i\mathcal{L}_\theta(\mathbf{x}_i) - (1-y_i)\log\left[1-e^{-\mathcal{L}_\theta(\mathbf{x}_i)}\right]. 
    \label{eq_03}
\end{align}
This maximum likelihood function represents a contrastive loss that attempts to minimize the reconstruction error $\mathcal{L}_\theta(\mathbf{x}_i)$ for fully frozen or normal $(y_i=1)$ time series while maximizing it for fully thawed or anomalous $(y_i=0)$ time series. The reason is that when $y_i=0$, $-\log\left[1-e^{-\mathcal{L}_\theta(\mathbf{x}_i)}\right]$ is monotonically decreasing as a function of the reconstruction loss $\mathcal{L}_\theta(\mathbf{x}_i)\ge0$. Thus, the minimization is equivalent to maximization of $\mathcal{L}_\theta(\mathbf{x}_i)$, providing a contrastive trade-off between the losses of normal and anomaly samples. In this representation, for the optimal parameters $\hat{\theta}$,  $p_{\hat{\theta}}(y_i= 1|x_i) = \eta_\theta(x_i)$ denotes the probability that each point $x_i$ belongs to a normal or frozen time series enabling a probabilistic retrieval. 

\subsection{Model Implementation and Training}
\label{sec:III.2}

The architecture of the FTC-Encoder is illustrated in Fig.~\ref{fig:02}. Both polarization channels of SMAP TBs and their differences (i.e., polarization signal) are used as inputs. The encoder block captures a low-dimensional latent space representation through a series of 1D convolution layers followed by ReLU activation functions and a dropout layer with 10\% rate. The decoder block upsamples the latent space through a series of 1D transposed convolution layers, the ReLU activation functions, and dropout layers. The filter size is one week but the number of filters varies from 32 to 64 through the encoder and decoder blocks.

For training, the SMAP TB time series from 2015 to 2017, with a minimum continuous length of one week, are utilized over the peak frozen winters (thawed summers) as normal (anomalous) points, using the ERA5 reanalysis data. To that end, the TB time series of frozen winters (thawed summers) are isolated when the ERA5 top soil and 2-m air temperatures are lower (greater) than 271 (275)~\si{K}. The safety margin of $\pm$2~\si{K} ensures that the selected segments represent frozen or thawed landscape conditions with a high probability, excluding shoulder seasons.  

We trained multiple FTC-Encoders for different land surface types and subgrid water fractions as shown in Fig.~\ref{fig:01}a to implicitly account for the interrelated landscape structural and radiometric properties. This training strategy aims to group pixels with analogous structural and radiometric properties to pave the way for improved FT-cycle retrievals. As shown in Fig.~\ref{fig:01}b and later on in Fig.~\ref{fig:03}, when the water fraction surpasses 5\% of the FOV, the sub-grid lake ice phenology significantly alters the TB time series. The model performance is tested for the SMAP TB time series from 2018 to 2020. Pixels containing ISMN ground-based stations are excluded from the training process to ensure a robust evaluation of the algorithm's performance across all years where SMAP and ISMN data overlap. 

\section{Results and Discussion}
\label{sec:IV}

In this section, the performance of the FTC-Encoder is quantified by common classification quality metrics obtained through comparisons with the available {\it in situ} observations of ground and air temperatures. We emphasize that satellite retrievals largely respond to the presence of liquid water in the soil-snow-vegetation continuum and not solely the soil temperature. Therefore, comparisons with ground-based observations may not be a perfect choice but perhaps the best practical one. 

The metrics are presented across various classes of land-cover types and sub-grid water fractions. Additionally, two annual FT)cycle retrievals with different land covers and water fractions are studied to further elaborate on the performance of the FTC-Encoder compared to the SMAP official product. Lastly, we elaborate on the spatial and temporal consistency of the FTC-Encoder, underscoring its performance in tracking probabilistic FT landscape states across Alaska's diverse land-cover types.

\subsection{Performance Statistics}

The FTC-Encoder produces a continuous measure $p(F)$, indicating the probability of frozen $\left(p(F)=1\right)$ or thawed $\left(p(F)=0\right)$ states of the landscape. A probability threshold of 0.5 is selected to determine the FT state in a binary setting for comparisons with SPL3FTP\_E product across the 34 ISMN ground stations (Fig.~\ref{fig:01}) -- reported through the confusion matrices in Table.~\ref{table:1}. The ground and air temperature measurements at the sites are used to determine the reference FT status of the landscape.

The confusion matrix quantifies the truly identified frozen states (TP: true positives), truly identified thawed states (TN: true negative), number of falsely identified thawed states (FN: false negative), and number of falsely identified frozen states (FP: false positive). Moreover, the classification performance is evaluated based on recall TP/(TP+FN), precision TP/(TP+FP), and accuracy (TP+TN)/(TP+TN+FP+FN) metrics for both frozen and thawed states.

\begin{table}[h]
    \centering
    \caption{Confusion matrix of FT binary retrievals from the FTC-Encoder and the SPL3FTP\_E (in parenthesis) based on the reference ISMN gauge observations of soil (a) and air (b) temperatures from 2015 to 2020. All values are presented as percentages with the accuracy reported in the lower-right cell.}
    \label{table:1}
    
{\bf (a) Ground Temperature} 
\vspace{3mm}
    
    \resizebox{\columnwidth}{!}{\begin{tabular}{c c|c|c|c|}
        \cline{3-5}
        & & \multicolumn{2}{c|}{\bf Actual Class} & \multirow{2}{*}{\bf Precision} \\ \cline{3-4}
        & & {\bf True (Frozen)} & {\bf False (Thawed)} & \\ \hline
        \multicolumn{1}{|c|}{\multirow{4}{*}{\centering \bf Predicted Class}} & \multirow{2}{*}{\bf T} & 46.2 & 2.8 & 94.3 \\
        \multicolumn{1}{|c|}{} & & \multicolumn{1}{m{1.5cm}|}{\centering (34.1)} & \multicolumn{1}{m{1.5cm}|}{\centering (2.4)} & \multicolumn{1}{m{1.5cm}|}{\centering (93.4)}  \\
        \cline{2-5} 
        \multicolumn{1}{|c|}{} & \multirow{2}{*}{\bf F} & 9.9 & 41.1 & 80.6 \\
        \multicolumn{1}{|c|}{} & & \multicolumn{1}{m{1.5cm}|}{\centering (22.0)} & \multicolumn{1}{m{1.5cm}|}{\centering (41.5)} & \multicolumn{1}{m{1.5cm}|}{\centering (65.3)}  \\
        \hline
        \multicolumn{2}{|c|}{\multirow{2}{*}{\centering \bf Recall}}  & 82.3 & 93.7 & 87.3 \\
        \multicolumn{2}{|c|}{} & (60.7) & (94.5) & (75.6) \\
        \hline
    \end{tabular}}
    
\vspace{3mm}
{\bf (b) Air Temperature}
\vspace{3mm}

    \resizebox{\columnwidth}{!}{\begin{tabular}{c c|c|c|c|}
        \cline{3-5}
        & & \multicolumn{2}{c|}{\bf Actual Class} & \multirow{2}{*}{\bf Precision} \\ \cline{3-4}
        & & {\bf True (Frozen)} & {\bf False (Thawed)} & \\ \hline
         \multicolumn{1}{|c|}{\multirow{4}{*}{\centering \bf Predicted Class}} & \multirow{2}{*}{\bf T} & 42.0 & 6.4 & 86.8 \\
        \multicolumn{1}{|c|}{} & & \multicolumn{1}{m{1.5cm}|}{\centering (32.4)} & \multicolumn{1}{m{1.5cm}|}{\centering (3.1)} & \multicolumn{1}{m{1.5cm}|}{\centering (91.3)}  \\
        \cline{2-5} 
        \multicolumn{1}{|c|}{} & \multirow{2}{*}{\bf F} & 5.9 & 45.7 & 88.5 \\
        \multicolumn{1}{|c|}{} & & \multicolumn{1}{m{1.5cm}|}{\centering (15.5)} & \multicolumn{1}{m{1.5cm}|}{\centering (49.0)} & \multicolumn{1}{m{1.5cm}|}{\centering (75.9)}  \\
        \hline
        \multicolumn{2}{|c|}{\multirow{2}{*}{\centering \bf Recall}}  & 87.6 & 87.7 & 87.7 \\
        \multicolumn{2}{|c|}{} & (67.6) & (94.1) & (81.4) \\
        \hline
        \end{tabular}}
\end{table}

The results for FT-cycle retrievals (Table~\ref{table:1}\,a) indicate improved performance compared to the SPL3FTP\_E across almost all classification metrics. When the soil temperature is used for FT labeling, the FTC-Encoder algorithm enhances the frozen state detection (TP) by approximately 12\%. The false detection of thawed state (FN) is also reduced by 12\%. These improvements manifest themselves in a marked increase of recall and accuracy by 21 and 15\%, compared to the SPL3FTP\_E product. The enhanced statistics can be mostly attributed to the reduction in falsely identified wintertime thawed states (FN), which can impair the results of SPL3FTP\_E perhaps due to the known effects of snow wetness on the NPR ratio as well as low differences in the NPR reference values \cite{derksen2017retrieving}.

When the FT labels are obtained based on the air temperature (Table~\ref{table:1}\,b), the true detection of the frozen state (TP) is improved by 9\%, while the incorrect detection of thawed states (FN) is dropped by 9.6\% compared to SPL3FTP\_E. At the same time, the misclassification of thawed states as frozen (FP) is 6.5\% for FTC-Encoder, which is larger than that of 3.1\% for the SPL3FTP\_E. This difference can be attributed largely to the instances where the air temperature is above zero, but the FTC-Encoder continues to produce frozen labels (i.e., $p(F)>0.5$). As is well understood, the soil temperature rises above the freezing point and drops below the freezing point later than the air temperature in the spring and fall respectively due to the latent heat release of soil thawing and freezing, known as the zero-curtain effect. Nevertheless, even in this labeling scenario, the TP rate and accuracy are increased by 20 and 5.6\%, respectively, compared to the SPL3FTP\_E.

\begin{figure*}
    \centering
    \includegraphics[width=0.9\textwidth]{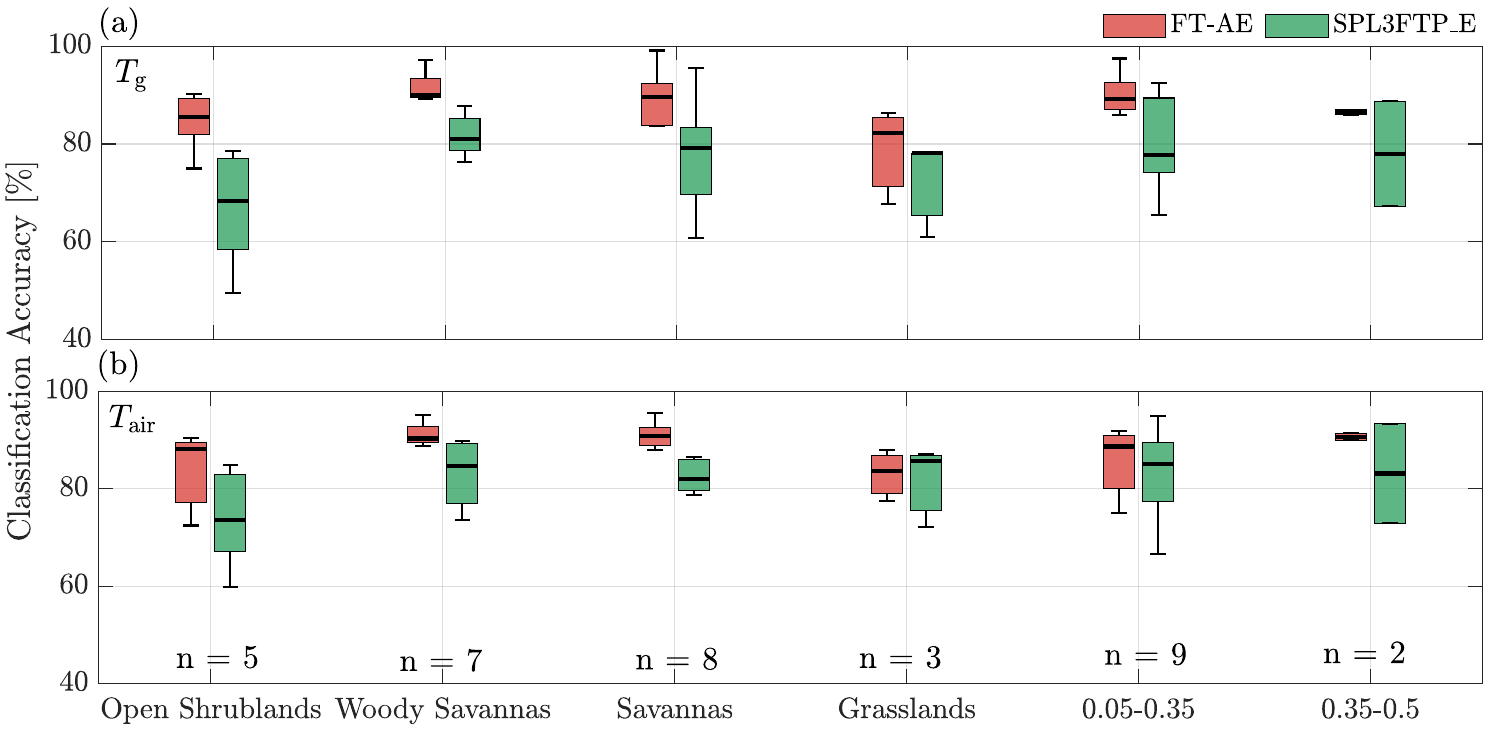}\\
    \caption{Freeze and thaw retrieval accuracy for the ISMN network sites (Fig.~\ref{fig:01}) grouped by land-cover classes and sub-pixel water fractions. The boxes represent the median and interquartile range while the whiskers extend the 5th and 95th percentiles.}
    \label{fig:03}
\end{figure*}

Fig.~\ref{fig:03} compares the performance of the FTC-Encoder and SPL3FTP\_E, by stratifying the accuracy metrics across different land-cover types and sub-grid water fractions for those SMAP pixels containing ISMN stations. When assessing the performance of the FTC-Encoder (SPL3FTP\_E) using soil temperatures, the mean accuracy across all land-cover types is approximately 90\% (70-80\%), showing a narrower interquartile range compared to SPL3FTP\_E. However, over grasslands, open shrublands, and the areas with high water fraction, the accuracy drops to around 83 (78), 85 (68)\%, and 87 (77)\%, respectively. A few insights can be offered to explain this reduction in accuracy. Firstly, the grasslands' and open shrublands' light vegetation and shallow snow cover provide minimum soil insulation, compared to other land-covers. Thus, the soil temperature can exhibit high temporal variability in response to changes in air temperature and an intermittent FT cycle is expected. This intermittency can decrease the accuracy, especially when a binary comparison is adopted through thresholding the FT probabilities. Secondly, when water fraction is high in satellite FOV, the TB time series cyclic structure primarily responds to ice in/out of the water bodies \cite{tikhonov2018theoretical} and wetness of the overlying snow cover, which can exhibit higher correlations with air temperatures rather than soil temperatures. These uncertainties are exacerbated considering that the reference currently rests on comparisons with a limited number of ISMN stations within the FOV.

When assessing the performance of the FTC-Encoder (SPL3FTP\_E) using air temperatures, the mean accuracy across all land-cover types is approximately 80--90 (75--85)\%. Thus, consistent with reported metrics in Tab.~\ref{table:1}, the accuracy does not change markedly for the FTC-Encoder but increases for SPL3FTP\_E, except in high water fraction areas. For instance, in Woody Savannas, the mean accuracy of FTC-Encoder remains the same around 89\%, whereas it rises by 3\% for SPL3FTP\_E. While, in pixels with high water fractions (0.35--0.5), FTC-Encoder's (SPL3FTP\_E) accuracy increases by 3 (4)\%. This observation indicates that perhaps the air temperature is a more realistic reference in explaining the FT cycle for those pixels with high water fractions, due to its higher correlation with the lake ice dynamics.

\begin{figure*}[b]
    \centering
    \includegraphics[width=1\textwidth]{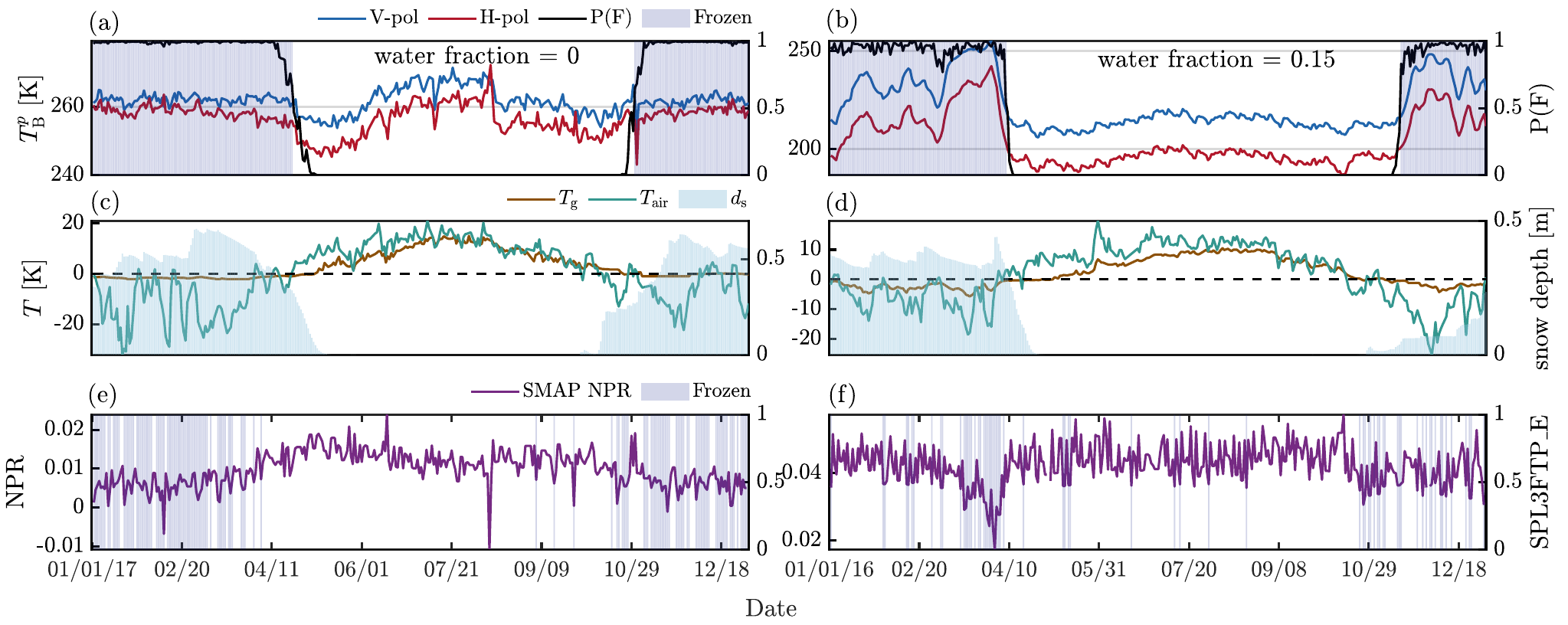}\\

    \caption{Time series of (a,b) SMAP TBs at vertical (V-pol) and horizontal (H-pol) polarization channels with the FTC-Encoder frozen probability $p(F)$, (c,d) in-situ air and soil temperatures from ISMN ground-based stations as well as ERA5 snow depth, and (e,f) the NPR ratio together with SPL3FTP\_E FT states over two SMAP pixels centered at (65.06\si{\degree}N, 146.71\si{\degree}W) and (63.92\si{\degree}N, 160.72\si{\degree}W) with sub-grid water fraction 0 (left column) and 15\% (right column).}
    \label{fig:04}
\end{figure*}

\subsection{Time Series Analysis}

Fig.~\ref{fig:04} compares the FT retrievals from the FTC-Encoder and SPL3FTP\_E based on ISMN air and soil temperatures at two SMAP pixels in Alaska with woody savannas land-cover and different water fractions. Consistent with our physical understanding, the L-band TBs remain warmer in winter than in summer. As soon as the air and soil temperature rise above the freezing point and with the onset of thawing, the surface emission suddenly decreases due to the presence of liquid water in the soil-snow-vegetation continuum resulting in a higher NPR ratio. The TBs gradually increase as the surface temperature increases and vegetation canopy becomes denser. As is evident, the presence of water bodies can change markedly the seasonality of the TB time series. 

In the case of a zero water fraction (Fig.~\ref{fig:04}a), the probability of a frozen state $p(F)$ is equal to 1 consistently throughout the winter when ISMN ground and air temperatures predominantly stay below the freezing point (Fig.~\ref{fig:04}c). In early spring, the probability gradually declines as the air temperatures oscillate around the freezing point, capturing the onset of the thawing season, manifested in the decline of ERA5 snow depth (Fig.~\ref{fig:04}c). In the middle of April, for four weeks, the $p(F)$ declines with some oscillatory behavior around 0.5 when air temperature varies above the freezing point over frozen ground. By early May, as the ground temperature rises above $0^\circ$~\si{C}, the probability of a frozen state is predominantly zero, indicating a fully thawed ground condition until early October. In mid-October, $p(F)$ sharply rises to 1 when air and ground temperatures drop below freezing over two weeks. It is worth noting that the spring thawing process, when $p(F)$ goes from 1 to 0, is approximately two weeks longer than the late fall freezing process, which is consistent with the known dynamics of the surface FT-cycle \cite{kim2018quantifying}. In early spring due to insulating snow cover and thus additional latent heat storage, the thawing period might take longer than the freezing time when the bare soils with no or a shallow snow cover become exposed to sub-zero air temperatures \cite{gao2022variability}.

It appears that the spikes of high NPR ratios (Fig.~\ref{fig:04}e) in the middle of the winter are coincident with sudden rises of air temperatures, leading to sporadic false detection of thawed states perhaps due to intermittent snow wetness rather than ground thawing. Since the reference NPR values for the frozen (0.007) and thawed states (0.012) are not significantly different in the SPL3FTP\_E, the discriminatory power of the algorithm can be reduced over snow-covered surfaces undergoing frequency surface melting and refreezing. However, due to the repeated applications of convolution operators and low-dimensional representation of the FT-cycle signal, the FTC-Encoder can be robust to the background noise leading to an accuracy of 89 (88)\% compared to 77 (79)\% for SPL3FTP\_E when reference ground (air) temperatures are used for labeling in this case.

A subgrid water fraction of around 15\% alters the cyclic structure of the TB time series (Fig.~\ref{fig:04}b) markedly. The depression due to the ground thawing reaches approximately 50~\si{K}, two times larger than that of the pixel with zero subgrid water fraction (Fig.~\ref{fig:04}a). As shown, the $p(F)$ dynamic is consistent with the ISMN ground temperature (Fig.~\ref{fig:04}b,d) indicating a continuous frozen state until the end of March. As the ISMN air temperature exceeds $0^\circ$\si{C} in early April over frozen ground, $p(F)$ declines rapidly. By mid-April, $p(F)$ approaches zero, when there is a significant drop in TBs nearly two weeks before the ground temperature rises above the freezing point in early May. This suggests that satellite retrievals may precede the onset of ground thawing when the sub-scale water fractional abundance is high. The reason can be attributed to a marked drop of the bulk surface emissivity due to the initial break up of the lake ice, even though the entire lake ice generally clears out after thawing of surrounding land surfaces \cite{higgins2021role}. 
 
During the freezing process, the ISMN air and ground temperatures drop gradually below the freezing point in late October, while $p(F)$ is still close to 0 and indicates a thawed state. After nearly one week of below-freezing ground temperatures in early November, a sharp increase in $p(F)$ values signals a completely frozen state by mid-November. Once again, this time lag is because the interior water bodies generally freeze later than the surrounding landscape and shift the content of the signal towards thawing. This suggests that the FTC-Encoder, based on its current training paradigm, produces a fully frozen state when the ground is frozen and the interior water bodies are fully covered by ice.

Compared to scenarios with zero water fraction (Fig.~\ref{fig:04}a), the FTC-Encoder's classification accuracy decreases (increases) by 2\% (1\%), while the SPL3FTP\_E shows a larger decrease of 16\% (13\%) when evaluated against in-situ ground-based soil and air temperature measurements, respectively. This decrease is chiefly due to frequent depressions of TB time series during the winter, resulting in large fluctuations of the NPR ratios and false detection of thawing periods (Fig.~\ref{fig:04}f). We conjecture that these fluctuations are the signatures of surface snow and ice melts in response to sporadic episodes of warm weather fronts as partially evidenced by ISMN air temperature data (Fig.~\ref{fig:04}d). Moreover, it appears that the SPL3FTP\_E algorithm also produces some sporadic false positives by detecting frozen states during the peak summer which can be attributed to the minimal difference between the freeze (0.038) and thaw (0.041) threshold references, rendering the classification sensitive to small but abrupt changes in the NPR values due to background effects of soil moisture and vegetation canopy.

\begin{figure*}
    \centering
    \textbf{\hspace{-1.0cm} Date \hspace{-0.1cm} 01/20 (A) \hspace{-0.1cm} 02/12 (B) \hspace{-0.1cm} 02-21 (C) \hspace{-0.1cm} 03/23 (D) \hspace{0.1cm} 04/10 (E) \hspace{0.1cm} 04/28 (F) \hspace{0.1cm} 05/23 (G) \hspace{0.1cm} 10/22 (H) \hspace{0.1cm} 11/18 (I) \hspace{0.1cm} 11/28 (J)}\par\medskip
    
    \frame{\includegraphics[width=0.08\textwidth,height=0.23\textwidth]{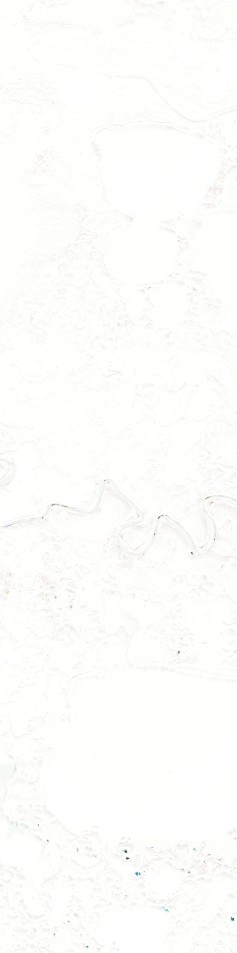}}\hspace{0.15cm}
    \frame{\includegraphics[width=0.08\textwidth,height=0.23\textwidth]{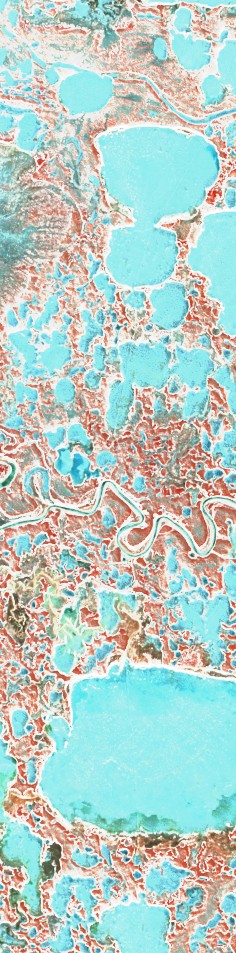}}\hspace{0.15cm}
    \frame{\includegraphics[width=0.08\textwidth,height=0.23\textwidth]{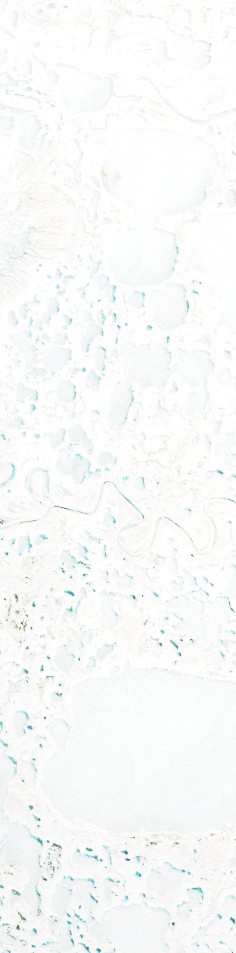}}\hspace{0.15cm}
    \frame{\includegraphics[width=0.08\textwidth,height=0.23\textwidth]{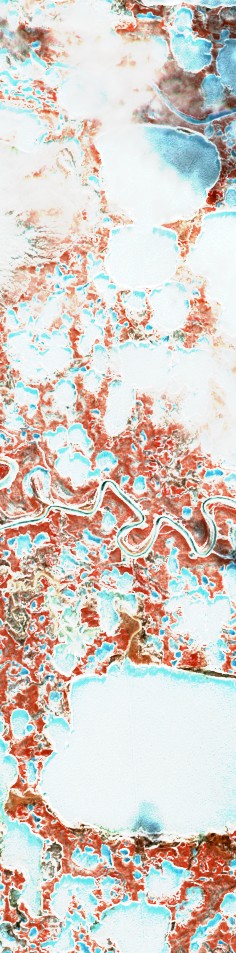}}\hspace{0.15cm}
    \frame{\includegraphics[width=0.08\textwidth,height=0.23\textwidth]{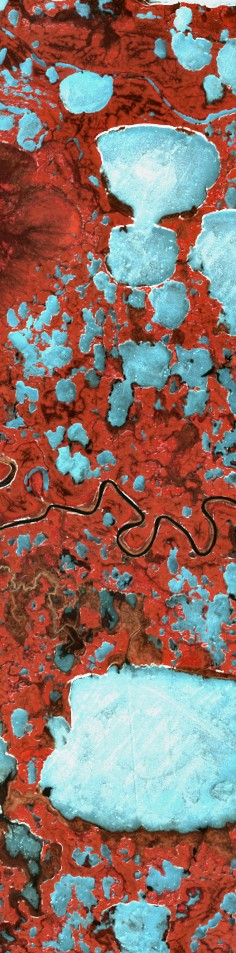}}\hspace{0.15cm}
    \frame{\includegraphics[width=0.08\textwidth,height=0.23\textwidth]{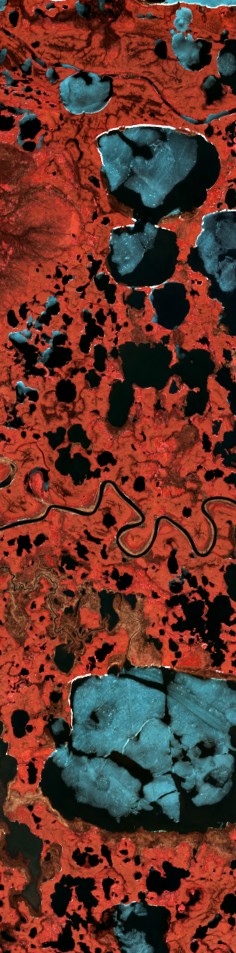}}\hspace{0.15cm}
    \frame{\includegraphics[width=0.08\textwidth,height=0.23\textwidth]{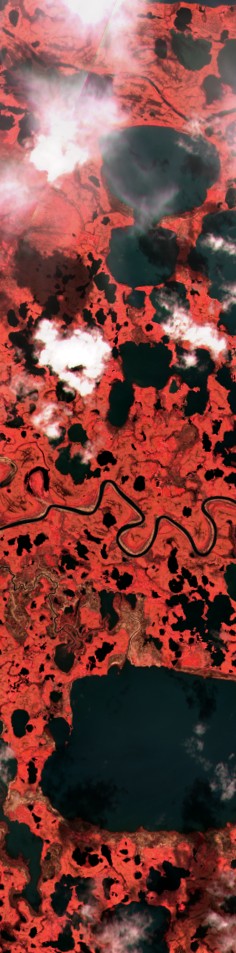}}\hspace{0.15cm}
    \frame{\includegraphics[width=0.08\textwidth,height=0.23\textwidth]{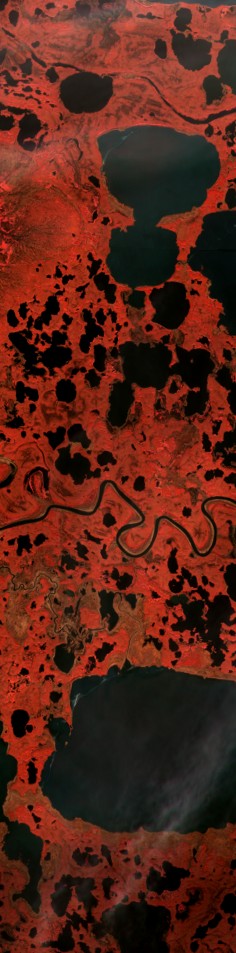}}\hspace{0.15cm}
    \frame{\includegraphics[width=0.08\textwidth,height=0.23\textwidth]{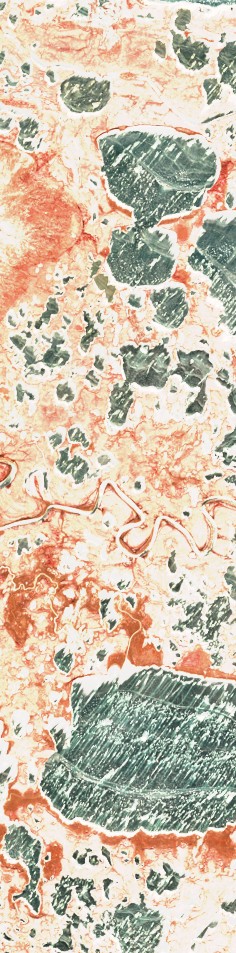}}\hspace{0.15cm}
    \frame{\includegraphics[width=0.08\textwidth,height=0.23\textwidth]{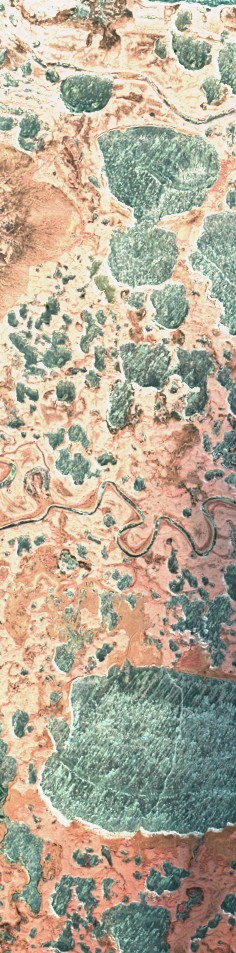}}\hspace{0.15cm}
    \\
    \vspace{3mm}
     \frame{\includegraphics[width=0.08\textwidth,height=0.23\textwidth]{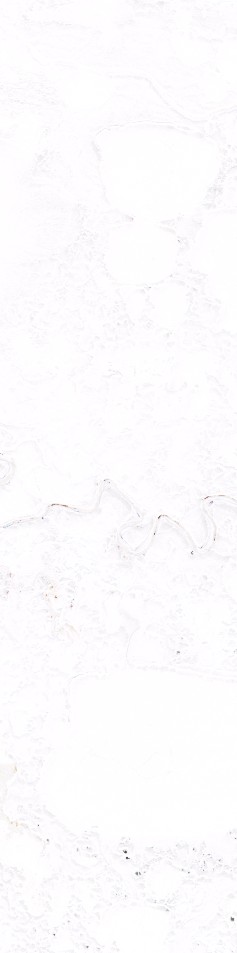}}\hspace{0.15cm}
    \frame{\includegraphics[width=0.08\textwidth,height=0.23\textwidth]{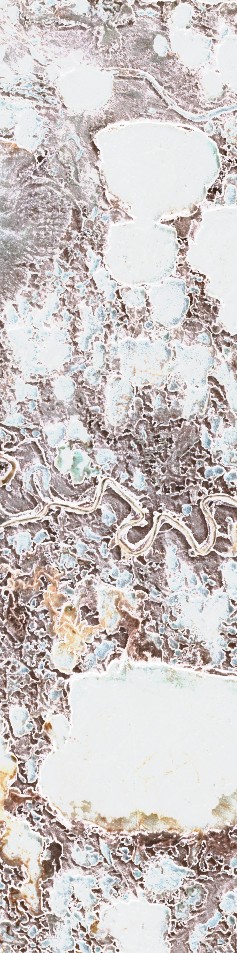}}\hspace{0.15cm}
    \frame{\includegraphics[width=0.08\textwidth,height=0.23\textwidth]{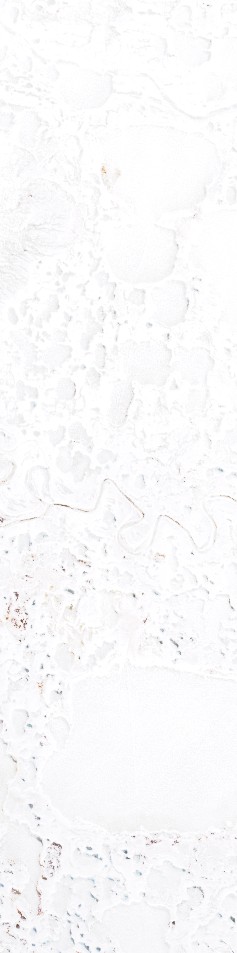}}\hspace{0.15cm}
    \frame{\includegraphics[width=0.08\textwidth,height=0.23\textwidth]{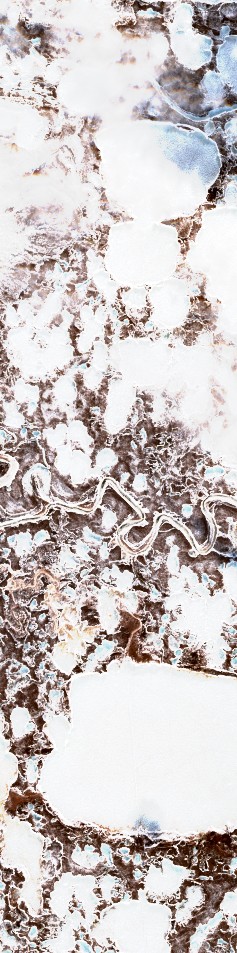}}\hspace{0.15cm}
    \frame{\includegraphics[width=0.08\textwidth,height=0.23\textwidth]{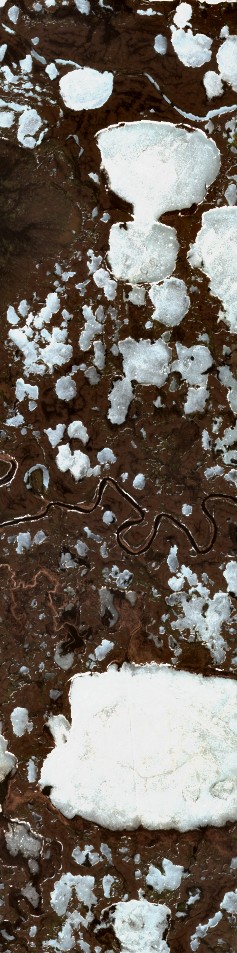}}\hspace{0.15cm}
    \frame{\includegraphics[width=0.08\textwidth,height=0.23\textwidth]{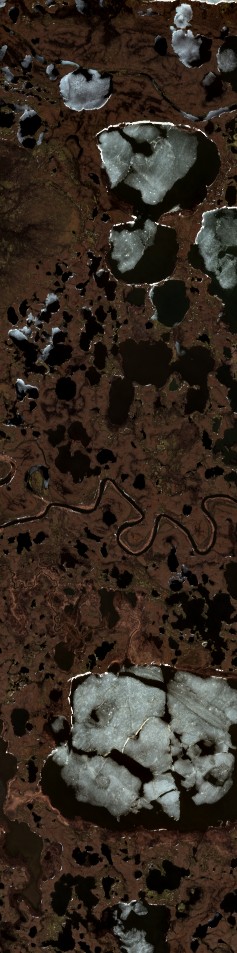}}\hspace{0.15cm}
    \frame{\includegraphics[width=0.08\textwidth,height=0.23\textwidth]{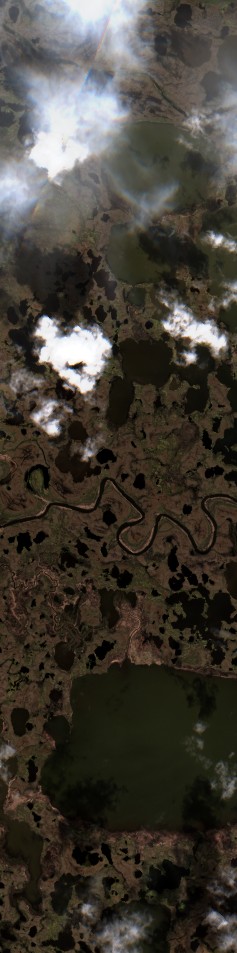}}\hspace{0.15cm}
    \frame{\includegraphics[width=0.08\textwidth,height=0.23\textwidth]{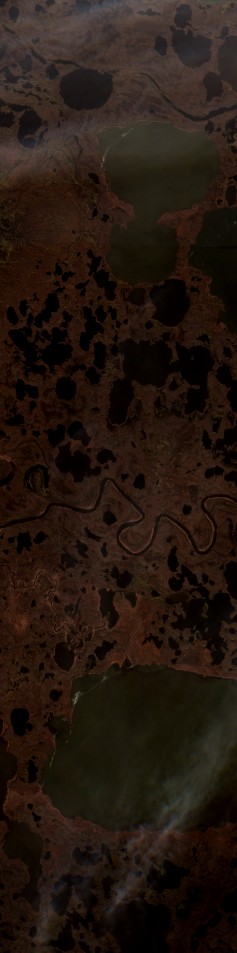}}\hspace{0.15cm}
    \frame{\includegraphics[width=0.08\textwidth,height=0.23\textwidth]{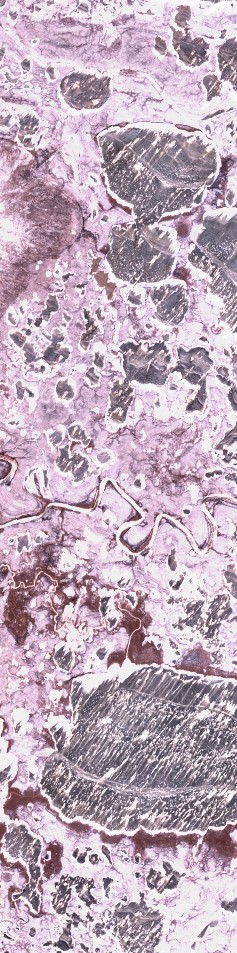}}\hspace{0.15cm}
    \frame{\includegraphics[width=0.08\textwidth,height=0.23\textwidth]{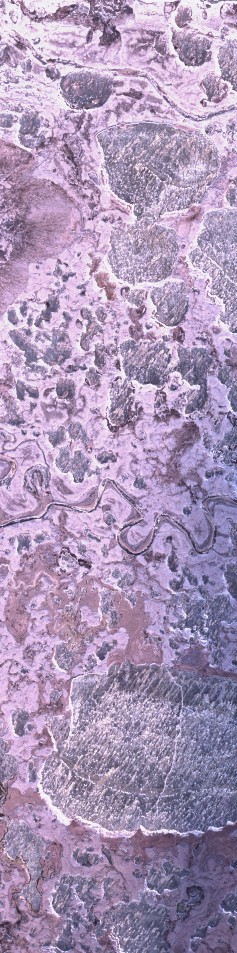}}\hspace{0.15cm}\\
    \vspace{2mm}
    \includegraphics[width=0.9\textwidth]{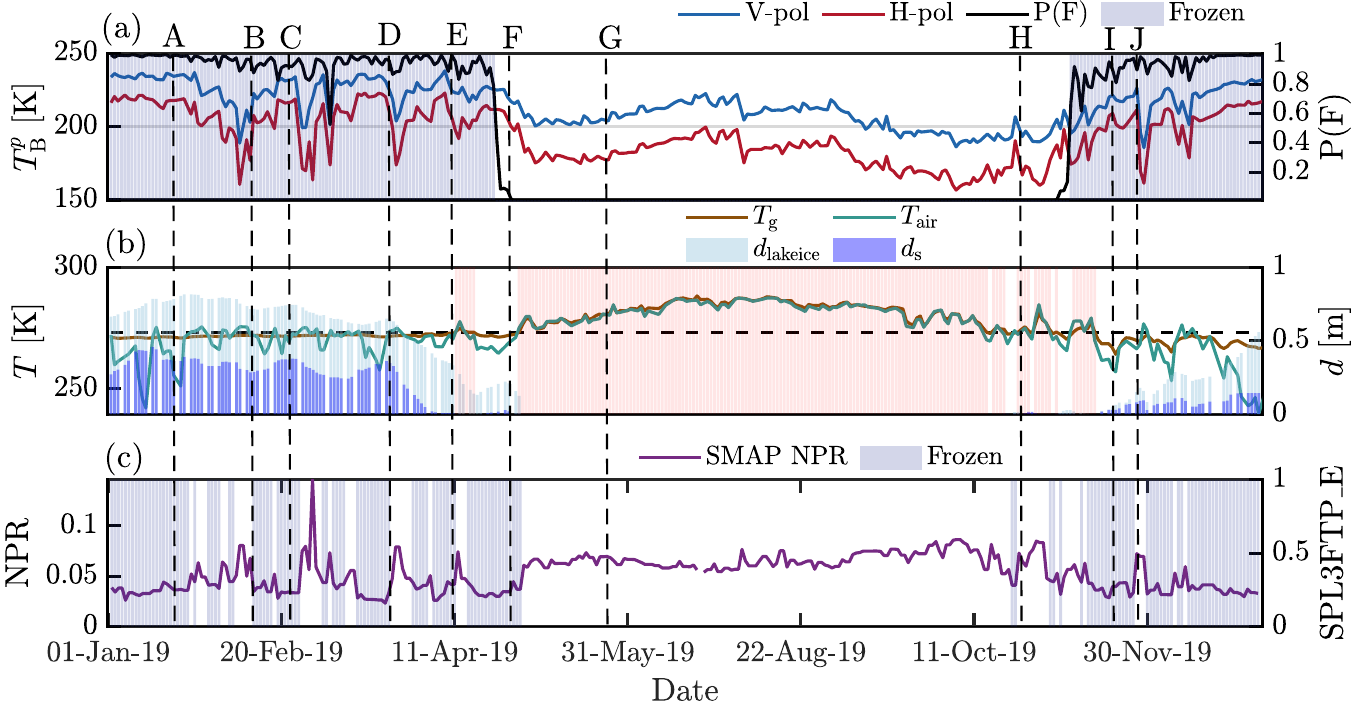} 
    
    \caption{The Sentinel-2 false composite and true color imageries over the SMAP footprint centered at 60.3398\si{\degree}N and 162.02\si{\degree}W in the calendar year 2019 (first and second rows, respectively), (a) the SMAP TBs at horizontal (H) and vertical (V) polarization with $p(F)$, (b) snow depth, lake ice depth, ground and air temperature measurements obtained from the ERA5 reanalysis dataset and (c) the NPR ratio with the SPL3FTP\_E freeze and thaw labels. The pink areas in panel (b) mark the ERA5 ground temperatures greater than 0\si{\degree C}. }
    \label{fig:05}
\end{figure*}

To further examine the impacts of the high water fractions on FT retrievals, we focus on SMAP overpasses in 2019 at latitude $60.34^\circ$N and longitude $162.02^\circ$W, with a water fraction of approximately 30\% (Fig.~\ref{fig:05}). In this pixel, there is no {\it in situ} ISMN station and we resort to reanalysis data and optical satellite imageries to explain the retrievals and expand the discussion. In this figure, the top rows show cloud-free snapshots from the 10-m Sentinel-2 composite band (first row) and true color (second row) imageries over the SMAP footprint.  

As expected, the TB time series are warmer in winter than in the summer. Unlike most pixels with zero water fraction, a marked drop of around 50~\si{K} is observed in TBs as the landscape transitions between FT states and the TB time series show higher temporal variability in winter than in summer. The ERA5 data support the hypothesis that these wintertime TB variabilities are due to the sporadic surface melting of snow or lake ice when the ground can remain predominantly frozen. When these intermittent melting episodes occur, the $p(F)$ by the FTC-Encoder drops by less than 30\% but remains largely above 0.5.  The FTC-Encoder is mostly robust in distinguishing these melting events from ground thawing based on the chosen probability threshold. On the other hand, the NPR-based approach seems capable of capturing these surface melting events beyond reanalysis data but at the same time can be vulnerable to some false labeling of these episodes as ground thawing.

\begin{figure*}
    \centering
    \includegraphics[width=1\textwidth]{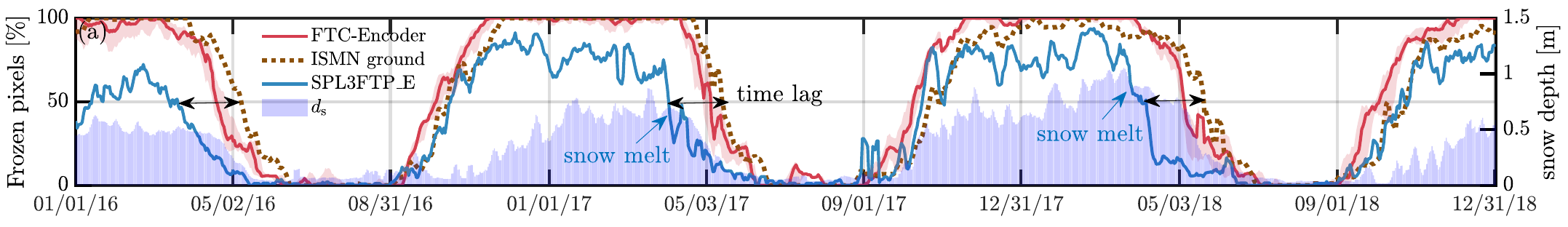}\\
    \includegraphics[width=1\textwidth]{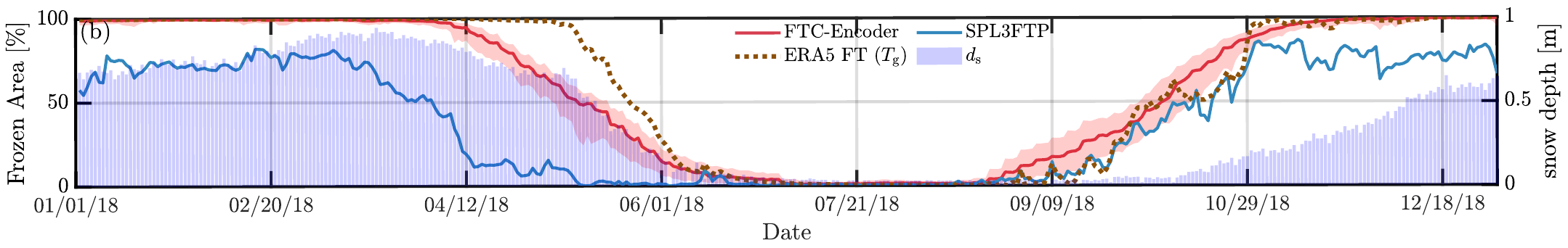}
    \caption{Percentage of (a) ISMN station and (b) grid cells classified as frozen over Alaska in the calendar year 2016-18 and 2018 by FTC-Encoder, SPL3FTP\_E product, and ground temperatures with average snow depth values obtained from ISMN station and ERA5 data. The shaded region for the FTC-Encoder represents the lower and upper bounds at the 0.05 and 0.95 probability levels. For ISMN observations (Fig.~\ref{fig:01}a) were selected when more than 15 gauge stations were available and the time series was smoothed out using a moving average window of 5 days for improved presentation. The pixels with a zero water fraction are considered when ERA5 data are shown to ensure a fair comparison.}
    \label{fig:06}
\end{figure*}

Close inspection of Fig.~\ref{fig:05} unravels the pros and cons of the two FT-cycle retrieval approaches. The temporal dynamics of TBs indicate a completely frozen state during January, consistent with the $p(F)$ by the FTC-encoder, SPL3FTP\_E FT state, and the Sentinel imagery on January 20 (point A). However, the retrieval discrepancies emerge in early February as the air temperature oscillates around the freezing point and the presence of liquid water within the FOV manifests itself as frequent depressions in the TB time series. A pronounced depression in the TB time series is apparent as the snow disappears from point January 20 (A) to February 12 (B), capturing the progress of a snowmelt event, as evidenced by the optical imageries. The TB values increase from February 12 (B) to February 21 (C) in response to another freezing episode as the FOV is again covered with fresh and dry snow as shown in the Sentinel imageries on February 21. The temporal dynamic of the TBs reveals another sequence of two FT cycles between points C and D. On April 10, while the surface is snow-free and large lakes are still covered with ice, the TBs begin a monotonically descending trajectory signaling the completion of the landscape thawing processes. Most small lakes are ice-free and ice floes are apparent in large lakes on April 28.

The FTC-Encoder retrieves a continuous frozen state from January 1 to April 23 as the $p(F)$ remains above 0.5. This is consistent with the ERA5 ground temperature except from April 11 to 17, when the ground temperature is above the freezing point, as shown with a pink shaded area in Fig.~\ref{fig:05}b. This time window is correctly labeled as a thawed state by the NPR thresholding; however, this approach identifies prior intermittent FT events that corroborate with the discussed snow accumulation and melting episodes based on the Sentinel imageries. For example, from February 4 to 14, a thawed state is detected based on the NPR ratio when the air temperature rises to 275~\si{K} for 1.5 weeks and the ground remains frozen. While the $p(F)$ drops by 5\%, the FTC-Encoder does not detect this event as a sporadic ground thaw event. The main reason is that NPR is sensitive to the instantaneous presence of water in the FOV and does not have any mechanism to reveal whether the signal emerges from soil, snow, vegetation, or lake ice. However, the FTC-Encoder differentiates the freeze and thaw signals through a series of temporal convolutional operations that can cancel transient melting signals shorter than its filter size (i.e., one week).   

The landscape freezing processes begin in late October. The ground temperature is predominantly below zero and covaries with the air temperature because of the weak thermal insulation of shallow snow covers. The Sentinel imageries on November 18 and December 28 show a frozen landscape with minimal snow cover and the lake surfaces are covered with new congelation ice. The ERA5 data represent frequent snow accumulation and melting/sublimation events with a depth of less than 25~\si{cm}. The ground temperature remains below zero after November 13, before the first sign of snow accumulation. The retrieved freezing probability sharply increases and exceeds 0.5 on November 9. The impacts of frequent surface snow melting and refreezing are apparent on $p(F)$ as it varies over time and gradually approaches one on November 18. The NPR ratio captures the freezing processes well and is more sensitive to sporadic snow-free ground freezing events than the FTC-Encoder. For example, the ERA5 temperatures and SMAP TBs record a short cold front that causes temporal landscape freezing for approximately 5 days in the first week of November. This freezing period is captured by SPL3FTP\_E but is missed by the FTC-encoder. As previously discussed, this sensitivity comes at the expense of some false detection and thawed states (i.e., November 17 and December 10).

\subsection{Spatiotemporal Assessments}

\subsubsection{Time series analyses}

Figure~\ref{fig:06}a compares the time series of pixel-level FT retrievals against the frozen percentage of the nearest ISMN station (see, Fig.~\ref{fig:01}) in calendar years 2016 to 2018. At the same time, the time series of FT retrievals in 2018 are evaluated (Fig.~\ref{fig:06}b) against the inferred FT cycle from ERA top soil temperature data over the entire study domain in Alaska. The results further highlight the fundamental differences between satellite retrievals of the FT cycle and those inferred from top-soil temperature data. The satellite retrievals lag behind temperature-based inferences of FT state during the thawing seasons but tend to follow a more consistent pattern during the freezing season.

As previously discussed, this time lag exists because the retrievals do not directly respond to changes in soil temperature but rather capture the presence of sufficiently extensive liquid water in the soil-snow-vegetation continuum and lake waters. Therefore, in snow-covered vegetated landscapes, satellite retrievals are susceptible to lagging behind the soil thawing in spring as the snowmelt begins earlier than the soil thaws. However, in late fall and winter, the retrieval and the surface soil FT cycle can be consistent when fresh, dry, and light snow cover does not interfere with the soil emission in the L-band. It is important to note that, a high sub-grid water fraction can widen this lag in spring (fall) as the ice-covered (ice-free) lakes are highly (negligibly) emissive and the breakup (formation) of ice often occurs after the thawing (freezing) of the surrounding land surfaces. 

The retrievals indicate that the time lag is shorter for the FTC-Encoder than that of SPL3FTP\_E in the thawing seasons and slightly longer in the freezing seasons. The results reveal that the FTC-Encoder is not fully agnostic to extensive and sustained snow-vegetation thawing despite evidence suggesting its robustness to transient above-surface melting depending on the choice of the probability threshold. An important observation is that, unlike ISMN and FTCENcoder data, the retrieved fraction of the frozen pixels (Fig.~\ref{fig:06}a) in the SPL3FTP\_E is always below 100\% throughout the year. 

\begin{figure*}
    \centering
    \includegraphics[width=0.80\textwidth]{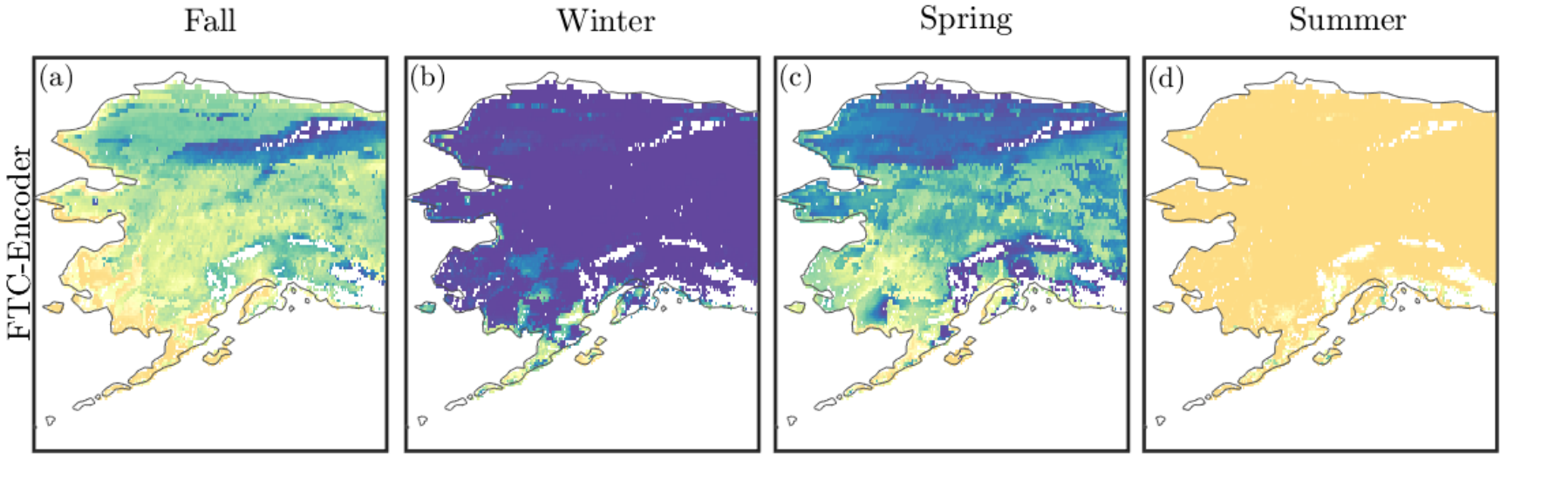}
    \includegraphics[width=0.80\textwidth]{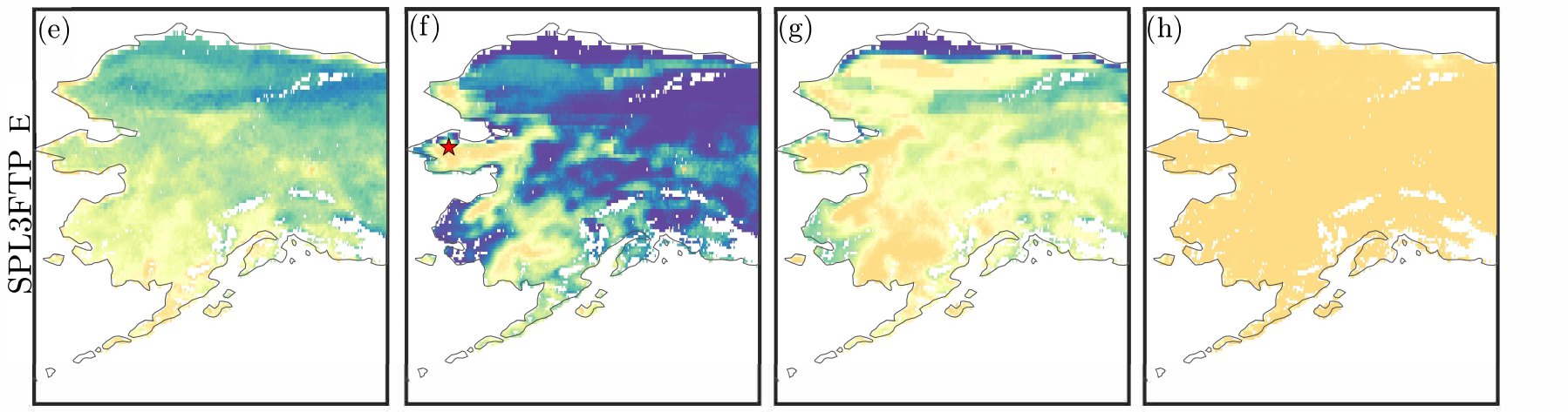}
    \includegraphics[width=0.80\textwidth]{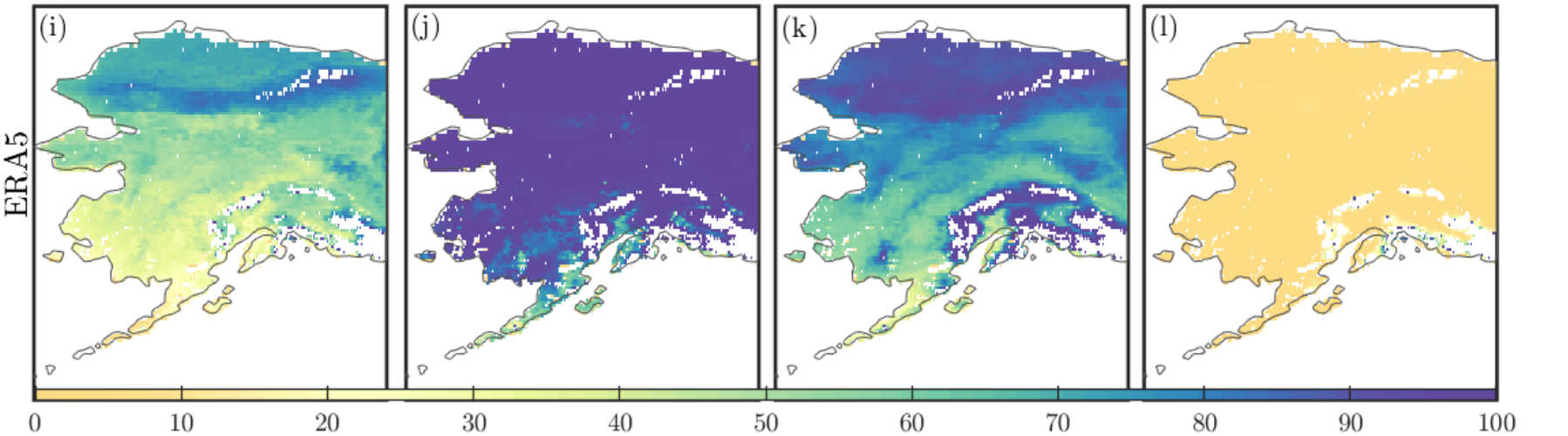}
    \caption{The seasonal percentage of frozen state retrievals using the FTC-Encoder (a--d), the SPL3FTP\_E product (e--h), and ERA5 FT state (i--l) using ground temperature data for Fall (Sep--Nov), Winter (Dec--Feb), Spring (Mar--May), and Summer (Jun--Aug) in calendar years 2018--2020.} \vspace{-5mm}
    \label{fig:07}
\end{figure*}

During the thawing season in calendar year 2017, more than half of the pixels were unfrozen on April 5 (Fig.~\ref{fig:06}a), almost 40 days earlier than the date inferred from ISMN data. This difference can be largely attributed to the snowpack melting, as evidenced by a noticeable decreasing trend in mean snow depth on that date. The FTC-Encoder is less sensitive to these initial snowmelt events as more than 50\% of the pixels were characterized as unfrozen on May 9, when extensive snowmelt occurred and the mean snow depth decreased below 15~\si{cm}. In the time series of 2018, snowmelt began around March 19 as the mean depth declined relatively sharply. The ground in more than half of the pixels became unfrozen on March 28 by the SPL3FTP\_E whereas such a date occurred on May 10 and 21 by the FTC-Encoder and ISMN gauge stations, respectively.

Figure~\ref{fig:06}b compares the fraction of the retrieved frozen pixels for all SMAP grids over Alaska in calendar year 2018 against the ERA5 counterpart -- inferred from the top-soil temperatures. The percentage of frozen pixels is consistent between ERA5 and FTC-Encoder during the peak winter before April 10. However, the SPL3FTP\_E only labels 75\% of the study domain as frozen during the peak winter. The satellite retrievals capture the onset of thawing on April 15 (FTCEncoder) and March 15 (SPL3FTP\_E) while the soil remains frozen until May 5 based on the ERA5 data.

We have already explained why the L-band FT estimates fundamentally differ from those inferred from the surface soil temperatures -- especially in spring. However, once again the results in Fig.~\ref{fig:06}b raise an important question: Why do the two satellite retrievals markedly differ concerning the retrieved frozen areas and timing of the landscape thawing? SPL3FTP\_E labels 75\% of pixels as frozen during the snow accumulation season in peak winter while all pixels are characterized as frozen by the FTC-Encoder before April 10. As will be shown later, most of these discrepancies are over the open shrublands, coastal, and mountainous regions. Moreover, the SPL3FTP\_E detects the onset of thawing, when the time series of frozen fraction decreases monotonically, 6 weeks earlier than that of FTC-Encoder. While the mean thawing onset inferred from FTC-Encoder retrievals is consistent with the initiation of mean snow depth reduction and thus a wide range of snowmelt across Alaska, the SPL3FTP\_E approach identifies the onset of melting when the snow depth time series is still increasing. 

\subsubsection{Spatial analyses}

The spatial pattern of the seasonal freezing probability elaborates on these discrepancies. Figure~\ref{fig:07} shows the seasonal mean of the frozen state probability derived from FTC-Encoder, SPL3FTP\_E, and ERA5 ground temperature data. This analysis covers three years of the test dataset from 2018 to 2020. Grid cells that remain unfilled represent pixels with glaciers, categorized under the permanent ice land-cover type.

Visual inspection indicates that the mean seasonal dynamics of the FT states are consistent between the FTC-Encoder and those inferred from ERA5 topsoil temperatures. However, the FTC-Encoder consistently shows a lower probability of frozen states in fall and spring compared to ERA5, particularly in the Far North and Southwest coastal regions. These areas are characterized by numerous lakes whose ice-cover dynamics affect the FT cycle of the landscape, leading to discrepancies between satellite retrievals and reanalysis of soil FT cycles. In the Southwest Region, especially over the Yukon Delta, the higher water fraction due to lakes and the dense network of rivers and streams are the primary causes of the observed differences, which can be as large as 20\%.

The SPL3FTP\_E retrievals (Fig.~\ref{fig:07}e-h) agree with ERA data during fall and summer but exhibit significantly lower freezing probabilities in winter and spring. The discrepancies are most evident in regions with extensive open shrubland biomes across upland tundra climate and mountain reliefs (e.g., the Brooks Range). For example, in winter (spring), SPL3FTP\_E produces a freezing probability of less than 30\% (10\%) over the Seward Peninsula and Southeast Alaska and around 40\% (20\%) in Interior Alaska and the Far North, where open shrublands dominate the type of the land cover (Fig.~\ref{fig:01}). However, the FTC-Encoder and ERA5 data show a significantly high freezing probability above 90\% (60\%) and 90\% (70\%) in winter (spring), respectively, across these areas. The freezing probability over the Brooks and Alaska Ranges is almost 40--50\% lower than the ERA5 counterpart in winter and spring. 

\begin{figure}
    \centering
    \includegraphics[width=0.5\textwidth]{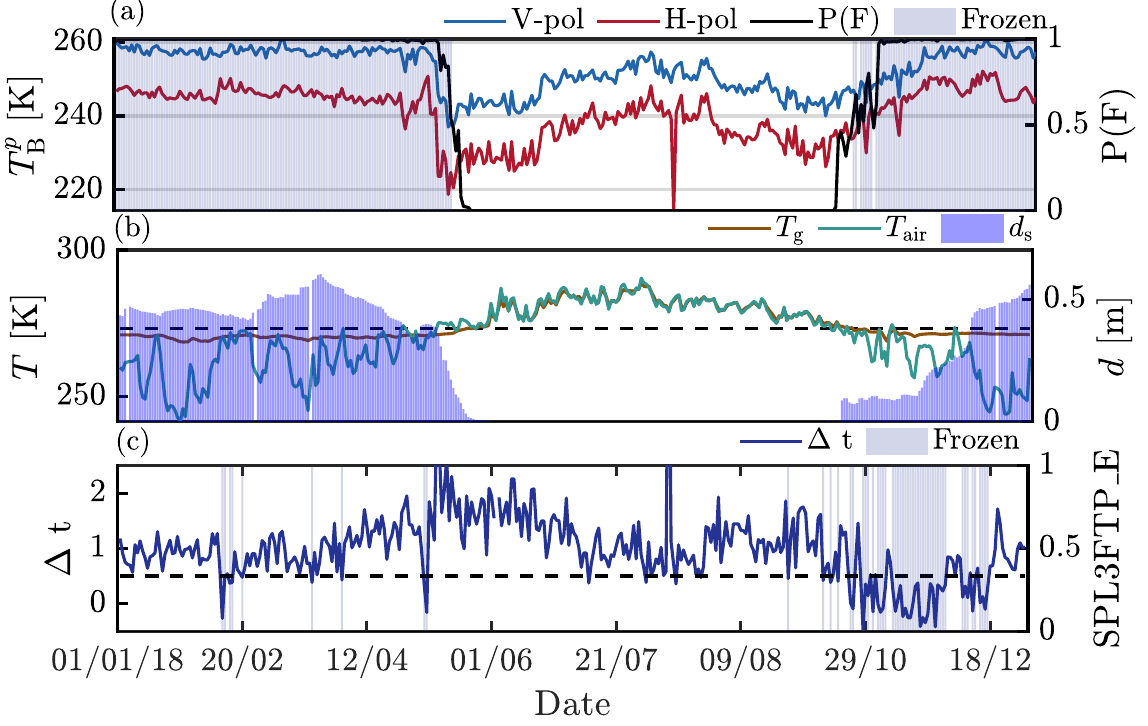}
    \caption{Time series of (a) SMAP TBs at vertical (V-pol) and horizontal (H-pol) polarization channels with the FTC-Encoder frozen probability $p(F)$, (b) the ERA5 air and ground temperatures as well as the snow depth, and (c) the seasonal thresholding computed using NPR ratio together with SPL3FTP\_E FT states over SMAP pixel centered at (66.01\,N, 164.60\,W) -- showed by a red asterisk marker on Fig.~\ref{fig:07}f.}
    \label{fig:08}
\end{figure}

These low regional freezing probabilities are the main reason for the observed underestimation of the frozen area fraction time series by SPL3FTP\_E in Fig.~\ref{fig:06}. The lack of skill for NPR thresholding over radiometrically rough surfaces has been previously studied and attributed to increased surface roughness and its depolarization impacts on the NPR values \cite{du2014classification,podest2014multisensor, kim2017extended,johnston2020comparing}. Fig~\ref{fig:08} shows the time series of the SMAP grid point in the Seaward Peninsula (marked in Fig.~\ref{fig:07}f) along with other necessary ERA5 ancillary data. The bottom panel shows $\Delta t = \frac{\rm{NPR(t) - NPR_{fr}}}{\rm{NPR_{th} - NPR_{fr}}}$, where NPR(t), NPR\textsubscript{th}, and NPR\textsubscript{fr} denote the NPR ratio at the time of interest, the NPR reference for thawed conditions, and the NPR reference for frozen conditions, obtained from SMAP official data sets.

As Fig.~\ref{fig:08} shows, the TB time series exhibits a typical seasonal pattern. The frozen probability computed by FTC-Encoder decreases around May 7, consistent with the above zero ERA5 air temperatures indicating the presence of water in the soil-snow-vegetation continuum. The frozen probability eventually decreases to zero on May 17 when the ground temperature rises above the freezing point. However, the SPL3FTP\_E indicates a thawed state for nearly the entire winter, from January to May, as $\Delta t$ remains greater than 0.5 most of the winter. This anomaly occurs because the difference between the thawed and frozen reference NPR values is small, making the $\Delta t$ relatively large and sensitive to small perturbations in the TB time series, which is the characteristics of all pixels over the open shrublands in lowland Arctic Alaska with tundra climate regimes. A small $\Delta t$ indicates that the climatology of surface emission polarization difference during the peak winter and summer is similar over this land cover type.

The vast majority of open shrublands across the Seward Peninsula, the western Brooks Range, the northwestern foothills, and the Arctic coastal plain of Alaska are populated by the perennial tussock tundra. The low and tall willow (Salix) shrubs are also found in floodplains and riparian corridors with canopy heights of more than 2~\si{m} in peak summer \cite{wells2022vegetation}. In the winter, the polarization difference in the surface emissivity increases drastically as the surface soil freezes and the short tussocks can be covered by dry snow. In the summer, as the snow melts the water content of ground and tussocks increases which can increase the polarization difference. However, the surface can be radiometrically rougher in the presence of green and densely spread tussocks with significantly higher vegetation water content than in winter which can decrease the polarization difference \cite{norouzi2011sensitivity}. Hence because of these competing effects, we conjecture that the effect of roughness, elevated vegetation water content, and soil moisture in summer can produce similar $\Delta_{\rm NPR}$ values in the winter. 


\begin{figure}
    \centering
    \includegraphics[width=0.35\textwidth]{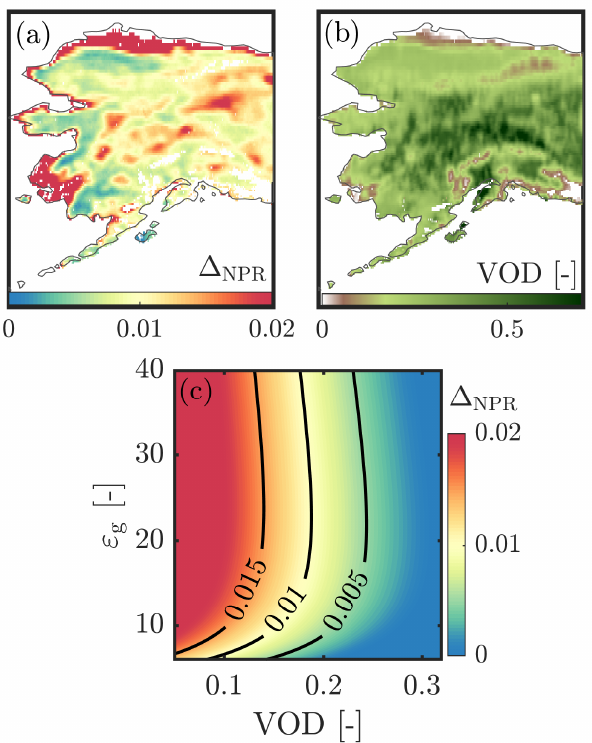}
    \caption{(a) The differences in NPR frozen and thawed reference values $\Delta_{\rm NPR}$, used in SPL3FTP\_E algorithm for seasonal thresholding, (b) mean summer VOD obtained from SMAP DCA algorithm in 2018 over the study region, and (c) experimental simulations of $\Delta_{\rm NPR}$ by the TO-snow emission model.}
    \label{fig:09}
\end{figure}

To support this conjecture, the difference between the NPR reference frozen and thawed values, $\Delta_{\rm NPR} = \rm{NPR_{th} - NPR_{fr}}$ is shown in Fig.\ref{fig:09}a. It can be seen that regions in Alaska where SPL3FTP\_E produces low freezing probability in winter (Fig.\ref{fig:07}f) are collocated with shrublands and $\Delta_{\rm NPR}<0.05$, characterized by mean summer vegetation optical depth (VOD) ranging between 0.2 and 0.4 (Fig.\ref{fig:09}b), and extremely low temperatures covered with tundra snow in winter, resulting in almost no emerging vegetation (not shown).

Figure~\ref{fig:09}c shows simulated $\Delta_{\rm NPR}$ using the TO-snow emission model \cite{kumawat2022passive, kumawat2023passive, kumawat2024global}. To that end, we compute $\rm NPR_{fr}$ by considering ground permittivity $\varepsilon_{\rm g} = 5$, snow density of 350~\si{kg.m^{-3}}, roughness coefficient $h = 1$, and vegetation single scattering albedo of 0.07, based on the ancillary parameters used in the SMAP Level 3 soil moisture product over open shrubland areas, with ground and vegetation temperatures of 270~\si{K}. For summertime simulations, all ancillary parameters are kept the same, and the snow-free ground and air temperatures are set to 280~\si{K}. The VOD and thawed ground permittivity are varied from 0 to 0.3 and 8 to 40 for $\Delta_{\rm NPR}$ computations.

As the figure shows, the simulated $\Delta_{\rm NPR}$ are largely sensitive to the changes in VOD and become less than 0.005 when VOD varies from 0.2 to 0.3, almost irrespective of the soil dielectric constant. This simulation supports the hypothesis that the diminished differences between $\rm NPR_{fr}$ and $\rm NPR_{th}$ are likely due to the effect of varying low vegetation (i.e., shrubs) than due to the changes in soil liquid water content. This suggests that the NPR approach can be sensitive to background noise over lightly vegetated surfaces and the applied threshold needs to be properly adjusted to alleviate the uncertainties. As demonstrated through several examples, the proposed deep learning approach is immune to these radiometrically complex scenarios and can properly learn the expected FT cycle over different land cover types.

\section{Conclusion}
\label{sec:V}

This study introduced the FTC-Encoder: a deep autoencoder neural network designed for probabilistic landscape FT-cycle retrievals using low-frequency microwave radiometry. The framework utilizes the time series of brightness temperatures (TB) at horizontal and vertical polarization to identify discriminative latent features of the FT state. The training relies on peak frozen and thawed segments from the TB time series, isolated conservatively using the reanalysis temperature data. The model learns the discriminative features of the FT state through a two-stage process, namely encoding (downsampling) and decoding (upsampling) of the selected segments of the TB time series using a contrastive loss function. This loss function is derived from the assumption that the likelihood of a time series belonging to a frozen or thawed state follows a Bernoulli distribution, thereby enabling the model to calculate the probabilities associated with the FT states. This framework obviates the need for precise ground-based labeling of FT states, an operational drawback for training machine learning models.

The results, derived from SMAP satellite observations over Alaska, demonstrate that the framework can achieve an accuracy of 87.3 (87.7)\%, using ISMN in-situ ground (air) temperatures to produce reference labels, respectively. It also reduces the false detection of thawed states by 12 (9.6)\% compared to the existing SMAP operational product. The analyses further elucidated the impacts of the sub-grid water fractions on the satellite FT retrievals of the landscape FT cycle. 

While L-band radiometry can sense the phase change of the water with the FOV, it seems imperative to pursue research to understand how to separate the signals of snowmelt and ground thawing through extensive field-scale experiments to deepen our radiometric and physical understanding of ground thaw and snowmelt processes. Retrieving ground permittivity \cite{kumawat2024global} can provide more direct information about the ground state; however, the retrieval problem becomes under-determined since ground and air temperatures must be assumed a priori, which influences the permittivity retrievals. Previous research \cite{li2022machine,zhong2022freeze,walker2022satellite,donahue2023deep} has attempted to address this problem by integrating higher microwave frequency information to distinguish between these signals, which can be incorporated into the FTC-Encoder. However, it is important to note that these approaches can be subject to significant uncertainties due to the low penetration depth of high-frequency channels through vegetation and wet snow.

Furthermore, it is important to emphasize that the FTC-Encoder was trained using distinct training data for various land-cover types and water-covered areas without prior knowledge of the reference climatological freeze-thaw states. Future research is essential to evaluate and enhance the capabilities of deep learning models for global-scale retrievals, particularly in Arctic landscapes characterized by high water fractions and organic permafrost soils. Additionally, there is a need to explore the connections between these outcomes and the carbon cycle in snow-covered permafrost and boreal forests. A demo code of FTC-Encoder is made publicly available at: \url{https://github.com/aebtehaj/FTCEncoder}.

\section*{Acknowledgment}
The research is mainly supported by grants from NASA's Remote Sensing Theory program (RST, 80NSSC20K1717) through Dr. Lucia Tsaoussi and the Interdisciplinary Research in Earth Science program (IDS, 80NSSC20K1294) through Dr. Will McCarty. The authors at the University of Minnesota also acknowledge the partial support from the Data Science Initiative (DSI) seed grants. A contribution to this work was made at the Jet Propulsion Laboratory, California Institute of Technology, under a contract with the National Aeronautics and Space Administration. The authors acknowledge the Minnesota Supercomputing Institute (MSI) at the University of Minnesota for providing resources (http://www.msi.umn.edu).

\bibliographystyle{IEEEtran}
\bibliography{Ref.bib}

\clearpage
\begin{IEEEbiography}[{\includegraphics[width=1in,height=2.25in,clip,keepaspectratio]{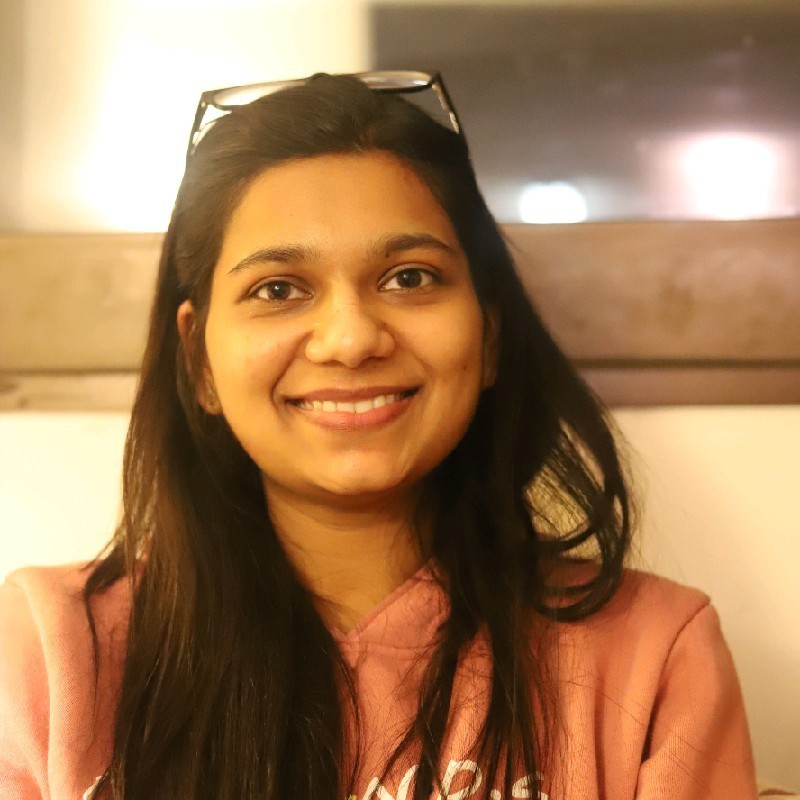}}]{Divya Kumawat}
received her B.Tech. in Civil Engineering from the Indian Institute of Technology (IIT), Roorkee, India in 2019. She is currently a PhD student in the Saint Anthony Fall Laboratory, the Civil, Environmental, and Geo-Engineering Department at the University of Minnesota -- Twin Cities, Minneapolis, MN, USA. Her research interests include microwave remote sensing of soil and snow, forward radiative transfer modeling, inverse problems, retrieval algorithms, and the applications of machine learning and deep Learning algorithms in the Earth Science domain.
\end{IEEEbiography}

\begin{IEEEbiography}[{\includegraphics[width=1in,height=1.25in,clip,keepaspectratio]{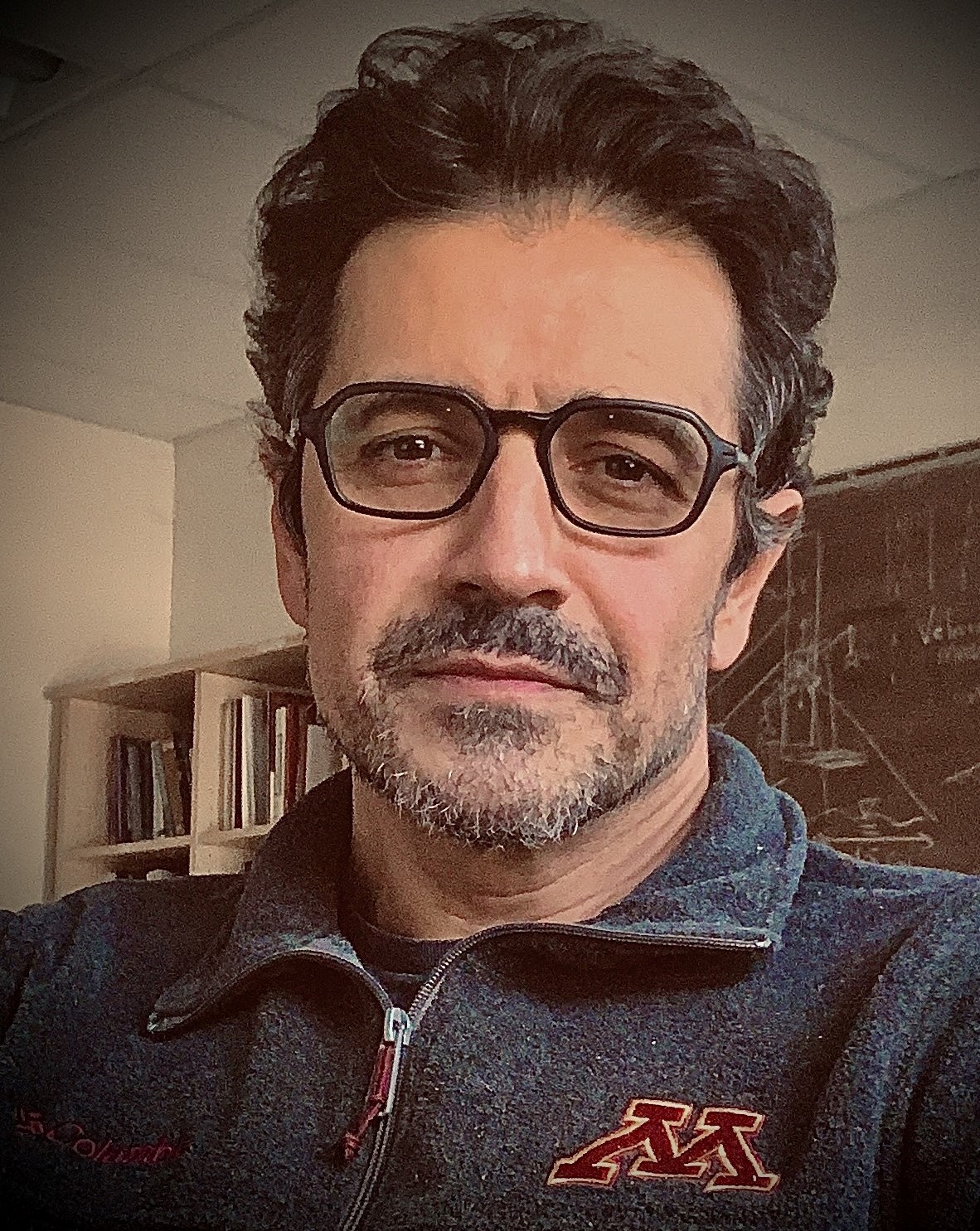}}]{Ardeshir Ebtehaj}
received his B.Sc. and M.Sc degrees in Civil and Environmental Engineering from the Iran University of Science and Technology, Tehran, Iran in 1999 and 2001. He also received an M.Sc. in Mathematics and a Ph.D. in hydrology from the University of Minnesota, Saint Anthony Falls Laboratory, Minneapolis, MN, USA in 2013. He is currently an associate professor in the Department of Civil, Environmental, and Geo-Engineering at the University of Minnesota -- Twin Cities. He served as a fellow of NASA's Earth and Space Science Program and is a recipient of NASA's Early Career New Investigator Award. 
\end{IEEEbiography}

\begin{IEEEbiography}[{\includegraphics[width=1in,height=1.25in,clip,keepaspectratio]{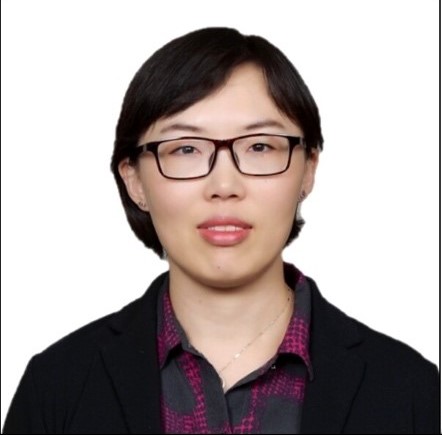}}]{Xiaolan Xu}
(Senior Member, IEEE) received her B.Eng. degree from Zhejiang University in China in 2006 and continued her academic journey at the University of Washington, Seattle, where she completed her M.S. and Ph.D. degrees in electrical engineering in 2008 and 2011, respectively. As a postdoctoral research associate, Dr. Xu joined the Jet Propulsion Laboratory (JPL) at the California Institute of Technology in Pasadena and later became a scientist in 2014. She received the URSI Santimay Basu Award in 2020. Her expertise revolves around applied electromagnetics, electromagnetic wave propagation, and scattering properties from snow-covered terrain, vegetated land surface, and bare soil. She specialized in the development of forward modeling and retrieval algorithms with a specific focus on their applications in Earth remote sensing from space. Dr. Xu's accomplishments underscore her commitment to advancing the field and her unwavering dedication to pioneering research.
\end{IEEEbiography}

\begin{IEEEbiography}[{\includegraphics[width=1in,height=1.25in,clip,keepaspectratio]{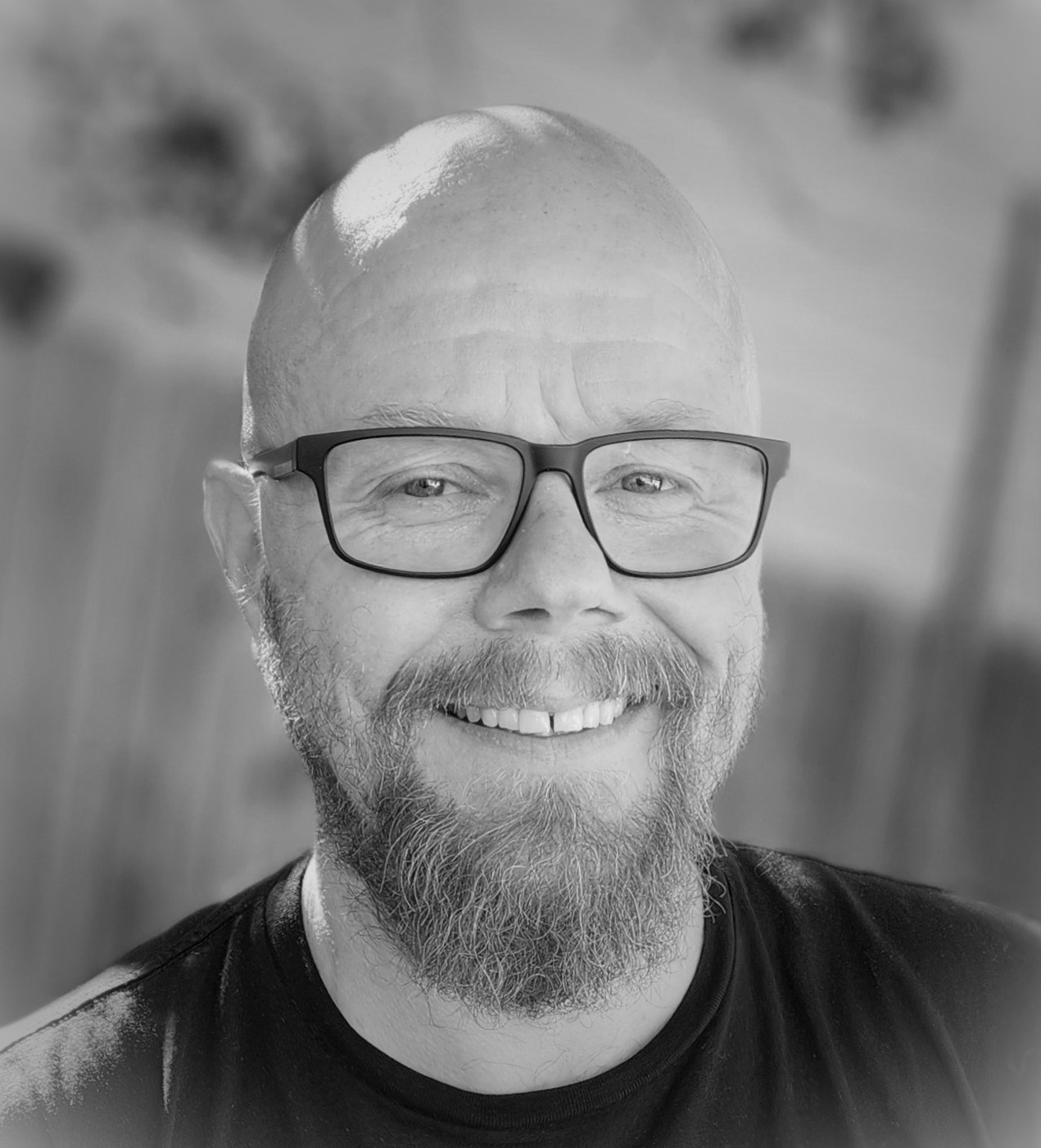}}]{Andreas Colliander}
(S’04-A’06-M’07-SM’08) received the M.Sc. (Tech.), Lic.Sc. (Tech.), and D.Sc. (Tech.) degrees in Electrical and Communications Engineering from Aalto University, Espoo, Finland, in 2002, 2005, and 2007, respectively. He is currently a Research Scientist with the Jet Propulsion Laboratory, California Institute of Technology, Pasadena. His research is focused on the development of microwave remote-sensing techniques. He is currently leading the calibration and validation of the geophysical retrievals of NASA’s SMAP mission and developing multi-frequency retrievals for ice sheets and polar atmospheres.
\end{IEEEbiography}

\begin{IEEEbiography}[{\includegraphics[width=1in,height=1.25in,clip,keepaspectratio]{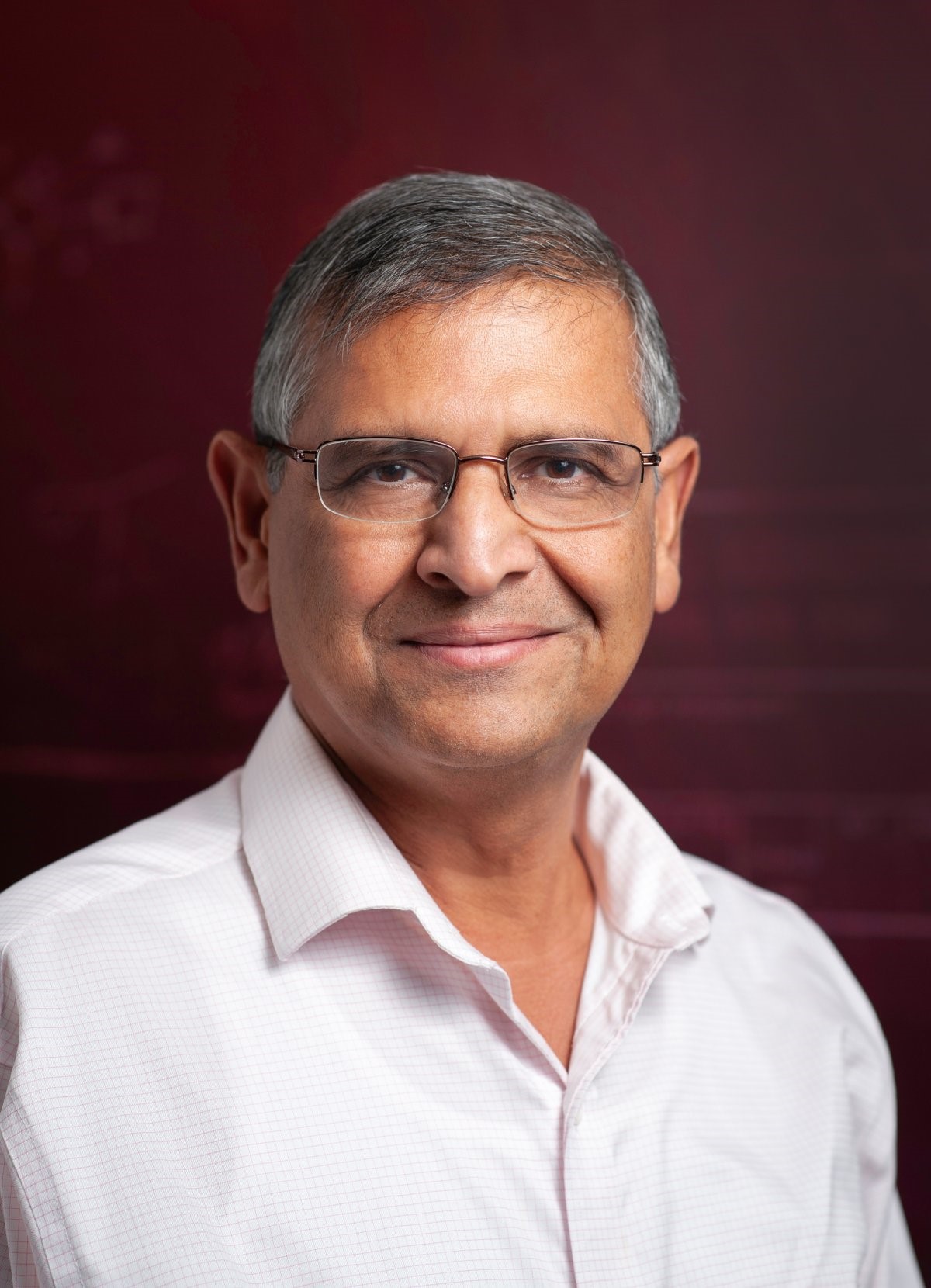}}]{Vipin Kumar}
is a Regents Professor of computer science at the University of Minnesota.  Kumar's research interests span data mining and high-performance computing (HPC) and their applications in Climate/Ecosystems and health care.  Kumar is a Fellow of AAAI, AAAS, ACM, IEEE, and SIAM.  His foundational research in data mining and HPC has been honored by the ACM SigKDD 2012 Innovation Award, the highest award for technical excellence in the field of Knowledge Discovery and Data Mining (KDD); and the 2016 Sidney Fernbach Award, one of IEEE Computer Society's highest awards in high-performance computing. 

\end{IEEEbiography}
\end{document}